\documentclass[journal]{IEEEtran}
\usepackage{microtype}
\usepackage{graphicx}
\usepackage{subcaption}
\usepackage{booktabs}
\usepackage{algorithm, algorithmic}
\usepackage[colorlinks, linkcolor=black, anchorcolor=black, citecolor=black]{hyperref}
\usepackage{stfloats}
\usepackage{amsmath}
\usepackage{amssymb}
\usepackage{mathtools}
\usepackage{amsthm}
\usepackage{bm}
\usepackage{multirow}
\usepackage{makecell}
\usepackage{caption}
\usepackage{graphicx}
\usepackage{float}
\usepackage{subcaption}
\usepackage[capitalize,noabbrev]{cleveref}
\usepackage{url}
\usepackage{cite}
\begin{document}

\title{Towards Fault Tolerance in Multi-Agent Reinforcement Learning}

\author{Yuchen Shi, Huaxin Pei, Liang Feng, Yi Zhang,~\IEEEmembership{Senior Member,~IEEE}, Danya Yao,~\IEEEmembership{Member,~IEEE}

\thanks{This work was supported in part by the National Natural Science Foundation of China under Grant 62133002. (\emph{Corresponding author: Huaxin Pei.})}
\thanks{Yuchen Shi is with the Department of Automation, Tsinghua University, Beijing 100084, China (e-mail: shiyuche21@mails.tsinghua.edu.cn).}%
\thanks{Huaxin Pei and Liang Feng are with QiYuan Lab, Beijing 100089, China (e-mail: peihuaxin@qiyuanlab.com; fengliang@qiyuanlab.com).}%
\thanks{Yi Zhang and Danya Yao are with the Department of Automation, Beijing National Research Center for Information Science and Technology (BNRist), Tsinghua University, Beijing 100084, China. (e-mail: zhyi@tsinghua.edu.cn; yaody@tsinghua.edu.cn).}%
}%

\maketitle
\begin{abstract}
	Agent faults pose a significant threat to the performance of multi-agent reinforcement learning (MARL) algorithms, introducing two key challenges. First, agents often struggle to extract critical information from the chaotic state space created by unexpected faults. Second, transitions recorded before and after faults in the replay buffer affect training unevenly, leading to a sample imbalance problem. To overcome these challenges, this paper enhances the fault tolerance of MARL by combining optimized model architecture with a tailored training data sampling strategy. Specifically, an attention mechanism is incorporated into the actor and critic networks to automatically detect faults and dynamically regulate the attention given to faulty agents. Additionally, a prioritization mechanism is introduced to selectively sample transitions critical to current training needs. To further support research in this area, we design and open-source a highly decoupled code platform for fault-tolerant MARL, aimed at improving the efficiency of studying related problems. Experimental results demonstrate the effectiveness of our method in handling various types of faults, faults occurring in any agent, and faults arising at random times.
\end{abstract}

\def\abstractname{Note to Practitioners}
\begin{abstract}
	Multi-agent systems based on MARL outperform those using traditional control methods in terms of performance but remain highly vulnerable to unexpected faults. To improve fault tolerance in such systems, we introduce an attention mechanism that enables the neural network to dynamically adjust its focus on fault-related information. Additionally, a prioritization sampling strategy is employed to select critical samples from collected experiences that are most relevant to current training needs. Experimental results across various fault types demonstrate significant improvements in fault tolerance, validating the robustness of our approach. These findings suggest that the proposed method has the potential to be applied to real-world scenarios, such as multi-robot systems and autonomous vehicle fleets.
\end{abstract}

\IEEEpeerreviewmaketitle

\section{Introduction}

\begin{figure*}[t]
	\begin{center}
	\centerline{\includegraphics[width=1\linewidth]{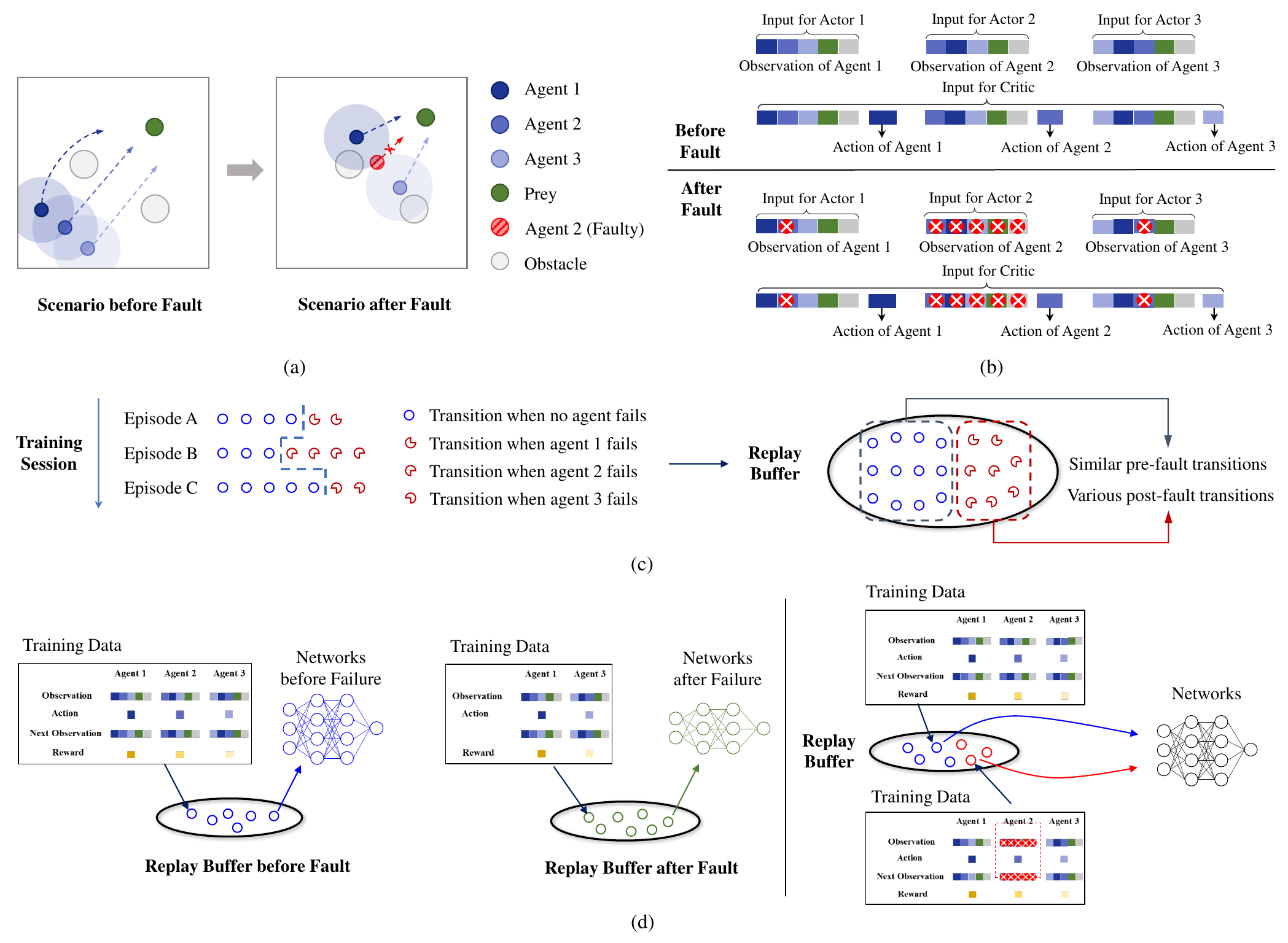}}
	\caption{(a) An illustration of a predator-prey system before and after the agent fault, where different shades of blue circles represent predators, and the green circle represents the prey. The large transparent circle around each agent represents its communication range. Following the fault of agent 2, which initially serves as a communication bridge, agent 1 and 3 suffer a loss of communication.
	(b) An illustration of the inputs for the actor and critic before and after fault. When agent 2 fails, its associated information becomes abnormal and is marked in red, leading to a disruption in the original input of the actor and critic networks.
	(c) An illustration of two natural ideas of handling faults. 
	The left part illustrates the idea of manually distinguishing the training data and the networks before and after the agent fault, 
	and the right part illustrates the idea of identifying the invalid information within the input automatically by the neural network.
	(d) An illustration of a replay buffer with transitions in 3 episodes. Blue circles represent similar pre-fault transitions and different imcomplete circles represent various post-fault transitions, reflecting imbalance of samples.}
	\label{introduction-figure}
	\end{center}
\end{figure*}

\IEEEPARstart{C}{ooperative} multi-agent systems have been widely applied in various fields, such as autonomous driving \cite{AV22,AV24,MARLAV24}, drone formation \cite{AUV19,AUV20}, and multi-robot control \cite{Robot23}. 
However, the occurrence of individual faults within the system is inevitable, especially in the complex operational environment, 
and it poses a significant threat to the system's ability to continue executing its intended tasks \cite{EffectComm19, ImpactAV20}. 
The existing studies demonstrate that there are numerous approaches to successfully enhance the fault tolerance of traditional cooperative control methods \cite{LRFT20,FT20}, 
such as the leader-following method \cite{Leadfollow23} and artificial potential field method \cite{Artipot21} in formation control, 
thereby reducing the impact of individual faults on the system's task execution.

Benefiting from the strong exploration ability of reinforcement learning in complex environments, 
multi-agent reinforcement learning (MARL) algorithms, 
employing the actor-critic \cite{PG99,10456565} based centralized training with decentralized execution (CTDE) framework \cite{MADDPG}, 
have emerged as the mainstream approaches for addressing cooperative problems in multi-agent systems. 
Notable examples include multi-agent deep deterministic policy gradient (MADDPG) \cite{MADDPG}, multi-agent proximal policy optimization (MAPPO) \cite{MAPPO}, etc. 
However, MARL faces unique challenges in dealing with unexpected agent faults within a system, as we discuss in this paper.

Taking the predator-prey system as an example, we present the primary challenges when an agent within the system fails. As shown in Fig. \ref{introduction-figure}(a), three agents with limited communication and perception ranges collaborate to pursue the prey, which employs a specific escape strategy. Due to the complex environment and challenging task, there is a distinct probability that an agent within the system may fail at any given time. When an agent fails, it becomes incapacitated, losing its ability to move, communicate, and perceive. This introduces several challenges, including disruptions in the communication structure and the need to reallocate tasks within the system. 

The impact of faults on MARL algorithms, both during training and execution, gives rise to two primary challenges. The first challenge is the chaotic inputs of networks, as illustrated in Fig. \ref{introduction-figure}(b). 
The sudden reduction of valid agents can lead to a decrease in the valid dimensions within the original state space, 
and the invalid information stemming from the failed agent can have detrimental effects on both the training and execution processes, significantly impacting the overall performance of the system. 
The second challenge is sample imbalance. To capture the unpredictability of fault timing and the affected agents, faults must be randomly assigned to different agents at random time steps during simulations. This randomness can lead to significant variability among the transitions stored in the replay buffer. As illustrated in Fig. \ref{introduction-figure}(c), in three episodes, different agents fail at different timesteps, resulting in the replay buffer storing many similar pre-fault transitions, while post-fault transitions can vary greatly. If uniform weights are assigned to transitions, inessential transitions may be repeated during training, thereby reducing training efficiency.

To tackle the aforementioned challenges, we propose fault-tolerant MARL solutions that integrate optimized model architecture with training data sampling strategy. The main contributions of our method are outlined below.

First, we introduce an attention-guided fault-tolerant method for MARL to tackle the chaotic state space, named \emph{\underline{F}ault-\underline{T}olerant Model with \underline{A}ttention on \underline{A}ctor and \underline{C}ritic}, AACFT for short. 
Before introducing AACFT, there are two natural remedial ideas to deal with faults. The first idea is to manually distinguish the training data and the networks before and after the agent fault (Fig. \ref{introduction-figure}(d) left). In such a case, experiences before and after faults are stored in different replay buffers, and actors and critics are separately set and trained. The drawback of this method is that it requires multiple replay buffers and actor-critic networks, which can be quite cumbersome. The second idea is to identify the invalid information within the input automatically by the neural network (Fig. \ref{introduction-figure}(d) right). In such a case, experiences before and after faults are stored in the only replay buffer, and a unique actor-critic network distinguishes whether a fault has occurred by the training data. Nevertheless, the presence of invalid information could lead to the learning of a suboptimal policy. Unlike the above two ideas, our proposed AACFT is capable of automatically identifying unexpected faults while appropriately tackling the special information within the chaotic state space. Specifically, we have carefully devised a method for configuring the input of the critic and actor networks and have integrated an attention module into the networks, building upon the MADDPG framework. In the critic, the observation of the faulty agent is no longer meaningful, and the attention module can then prioritize shifting attention away from the observations of the faulty agent and focus on other relevant information within the input. Within the actor, the observation of each agent encompasses the states of other agents, enabling the recording of the state when a fault occurs. The attention module can then dynamically modulate the level of attention assigned to the fault information based on its impact on the system.

Second, we employ a reweighting method that prioritizes the sampling of more challenging transitions for the current model, effectively mitigating the sample imbalance issue. Specifically, we extend the Prioritized Experience Replay (PER) method \cite{PER16} by creating separate priority queues for each actor and critic, with priorities determined by their respective losses. This allows each module to focus on learning from the transitions that are most relevant to it at any given time. Typically, tasks become more complex and diverse after a fault, increasing the difficulty of training. After initial training, when agents have developed a basic pre-fault strategy, the importance of training on post-fault transitions becomes even greater. In this context, our method effectively mitigates sample imbalance by prioritizing critical transitions.

Third, we conduct a comprehensive comparison of experimental results across various baselines and scenarios, demonstrating the effectiveness of our method in addressing potential faults. We provide visualizations of the attention mechanism, highlighting the dynamic changes in attention allocation within the critic and actor before and after a fault occurs. Additionally, we validate the prioritization mechanism through comparative experiments and analyze shifts in the distribution of sampled transitions. Furthermore, we evaluate our method's performance both under fault-free conditions and at different time points when faults occur, confirming its ability to seamlessly handle faults at any moment.

To support researchers in studying fault tolerance in MARL, we design and open-source a new code platform, along with our proposed algorithm\footnote{\url{https://github.com/xbgit/FaultTolerance_AACFT}}. Beyond the general functionalities of typical MARL platforms, our platform specifically addresses the impact of faults on various components, including the environment and the algorithm. By decoupling faults from other modules, researchers can easily customize and experiment with faults, algorithms, environments, and more, facilitating modular comparative studies.

The rest of the paper is organized as follows. In Section \ref{section2}, we introduce the related works, followed by background knowledge in Section \ref{Background}. Our approach is described in Section \ref{Methods}. In Section \ref{platform}, we introduce the framework and features of our open-sourced platform. In Section \ref{Experiments}, we show the experimental results. We conclude finally in Section \ref{conclusion}.

\section{Related Works}
\label{section2}
\subsection{Fault tolerance}
In this paper, we focus on addressing highly critical system-level faults in agents, where the affected agent loses its ability to observe or act, as opposed to faults occurring in specific components of the agent \cite{ft00, ft16,9400475,10005174}. In the existing studies, Pei \emph{et al.} \cite{FTtrans22} construct a rule-based model to achieve fault tolerance in multi-vehicle cooperative driving at signal-free intersections. Kamel \emph{et al.} \cite{FTCC18} design task-reassignment algorithms taking use of Hungarian algorithm, to ensure the completion of robot teams task after a robot's fault. Nevertheless, these works do not specifically aim to achieve fault tolerance in MARL algorithm. Some researchers focus on robust MARL \cite{RMARL19, RMARL23}, emphasizing performance balance when agent actions deviate. However, our research concentrates on enhancing performance in the event of specific types of faults, which is not the strength of robust MARL.
\subsection{MARL}
CTDE framework for MARL, including QMIX \cite{QMIX}, MADDPG \cite{MADDPG}, COMA \cite{COMA}, MAPPO \cite{MAPPO}, 
avoids the problem of non-stationarity when agents learn independently, 
and also avoids the problem of action space dimension explosion when a multi-agent system is modeled as a single agent.
Recently, many studies have endeavored to enhance MARL algorithms, aiming to increase their adaptability to  real-world MARL environments \cite{AttEvolveMARL19,MARL_TASE1,Asy23,Race23,MARL_TASE2,10759737}, 
such as R-MADDPG for partial observable environments \cite{RMADDPG20}, 
and MADDPG-M for environments with extremely noisy observations \cite{MADDPGM18}.
Similarly, we propose our model based on MADDPG, making it capable of fault tolerance problems in MARL. 
\subsection{Attention}
Attention mechanism has become a popular technique due to its superior performance, 
interpretability, and ease of integration with basic models \cite{AttLR21}. 
Attention mechanism is widely employed for a variety of deep learning models across many different domains and tasks, 
including natural language processing \cite{Translation15, attention17, AttLLM24}, 
computer vision \cite{RecurAtt14, VIT}, reinforcement learning \cite{MAAC, AttRecurrMARL23,AttGuide23, Att_TASE}.
Iqbal \& Sha \cite{MAAC} using centrally computed critics that share an attention mechanism, 
dynamically selecting which agents to attend to, enable more effective learning in MARL. In our approach, the attention mechanism plays a crucial role in enhancing fault tolerance.

\subsection{Transitions Reweighting}
Transitions reweighting is an effective method to improve utilization efficiency of transitions. There are typically two types: prioritizing easy transitions and prioritizing difficult transitions. The former is suitable for addressing issues with noisy labels in the dataset, as it tends to favor lower-loss examples which are more likely to be clean data \cite{CL17}. It is also suitable for solving extremely difficult problems that require a step-by-step approach \cite{CL23}. The latter prioritizes higher-loss samples that are more likely to belong to minority cases, thereby alleviating sample imbalance, such as Hard Example Mining (HEM) \cite{HEM16} and Prioritized Experience Replay (PER) \cite{PER16}. Considering that data in RL problems is clean and there is sample imbalance in fault tolerance problems, we extend PER to fault tolerance problems based on MARL and adapt it to our proposed AACFT.

\section{Background}
\label{Background}
\subsection{Markov Games}
A multi-agent system with $N$ potentially faulty agents can be modeled as decentralized partially observable Markov decision processes (Dec-POMDPs) \cite{Markov94, POMDP98},
defined as $<S,A,T,R,Y,O,\gamma,F>$, where ${S}$ is the set of states, ${A=\{A_1,...,A_N\}}$ is the set of joint actions, 
${T}$ is the transition function, ${O=\{O_1,...,O_N\}}$ is the set of observations, 
and ${Y=\{Y_1,...,Y_N\}}$ is the set of observation functions. 
At each time step $t$, agent ${i}$ receives a partial observation ${o_{i}=Y_i(s):S\rightarrow O_i}$, 
takes action ${a_{i}}$ according to policy ${\pi_{\theta_i}}$: ${a_{i}=\pi_{\theta_i}(o_{i}):O_i\rightarrow A_i}$, 
and receives a reward $r_i=R(s,a_1,...,a_N):S\times A_1 \times ... \times A_N \rightarrow \mathbb{R}$. 
Assume that fault occurs under a certain probability ${p}$, which is defined by state ${s}$ and the current time ${t}$: ${p=F(s,t)}$. 
Therefore, the environment at the next time step can be described as ${s'=T(s,a_{1},...,a_{N},p)}:S\times A_1\times ...\times A_N \rightarrow S$. 
The objective for each agent is to maximize its expected discounted reward ${\mathbb{E}[R_i]=\mathbb{E}[\sum_{t'=t}^{T}\gamma^{t'-t}r_{i,t'}]}$, 
where ${\gamma\in[0,1]}$ is the discount factor, $r_{i,t'}$ is the reward at time step $t'$ and $T$ is the end time step of an episode.

\subsection{Multi-Agent Deep Deterministic Policy Gradient (MADDPG)}
MADDPG \cite{MADDPG} is a variant of the deterministic policy gradient algorithm \cite{DDPG16} for multi-agent systems, 
with centralized Q-value function and decentralized policies.
Considering $N$ agents with continuous deterministic policies $\mu=\{\mu_{\theta_1},...,\mu_{\theta_N}\}$, parameterized by $\theta_i$,
the policy gradient for agent $i$ is:
\begin{equation}
	{\nabla_{\theta_i}J(\theta_i)}=\mathbb{E}_{U(\mathcal{D})}[\nabla_{\theta_i}\mu_{\theta_i}(a_i|o_i)\nabla_{a_i}Q_i^{\phi_i}(x,a)|_{a_i=\mu_{\theta_i}(o_i)}].
\end{equation}
$Q_i^{\phi_i}(x,a)$ is agent $i$'s Q-value function, parameterized by $\phi_i$, that accepts all agents' actions $a=[a_1,...,a_N]$ and state information $x$. 
In this paper, $x$ is defined as the concatenation of all agents' observations $[o_1,...,o_N]$.
$U(\mathcal{D})$ denotes experiences sampled from the replay buffer $\mathcal{D}$,
which stores tuples $(x,x',a,r_1,...,r_N)$, encapsulating all agents' experiences.
$x'$ signifies the next state following the environment transition.
$\phi_i$ is updated via:
\begin{equation}
	\begin{split}
		\mathcal{L}(\phi_i)=\mathbb{E}_{U(\mathcal{D})}[(Q_i^{{\phi_i}}(x,a)-y_i)^2],\\
		y_i=r_i+\gamma Q_i^{\phi'_i}(x',a'_1,...,a'_N)|_{a'_j=\mu_{\theta'_j}(o_j)}.
	\end{split}
\end{equation}
$\mu'=\{\mu_{\theta'_1},...,\mu_{\theta'_N}\}$ is the set of target policies with delayed parameters $\theta'_i$.

Notablely, $Q_i^{\phi_i}(x,a)$ is referred to as a critic, taking in all agents' observations and actions during centralized training.
$\mu_{\theta_i}(o_i)$ is referred to as an actor, taking its local partial observation and output the action decentralizedly. 

\subsection{Attention}
Suppose we have $n$ embeddings $\{e_1,...,e_n\}$ corresponding to $n$ encoded pieces of information. 
The attention weight from $e_i$ to $e_j$, presented as $\alpha_{ij}$, is calculated 
based on the similarity between the query of $e_i$ and the key of $e_j$. 
A normalization method follows, such as Softmax:
\begin{equation}
	\alpha_{ij}=\frac{\exp((W_q e_i)^T(W_k e_j))}{\sum_{r=1}^{n}\exp((W_q e_i)^T(W_k e_r))},
	\label{att1}
\end{equation}
where $W_q$ is query martix that transforms $e_i$ into a query, and $W_k$ is key martix that transforms $e_j$ into a key.
Then, the attention embedding $b_i$ for $i$-th piece of information is the weighted sum of the values for $\{e_1,...,e_n\}$:
\begin{equation}
	b_i=\sum_{j=1}^{n}\alpha_{ij} W_v e_j,
	\label{att2}
\end{equation}
where $W_v$ is value martix that transforms $e_j$ into a value. 
The resulting $b_i$, synthesized from $e_1,...,e_n$ with respective weights, can then be utilized for the subsequent information extraction.

\begin{figure*}[h]
	\begin{center}
	\centering
	\includegraphics[width=0.95\linewidth]{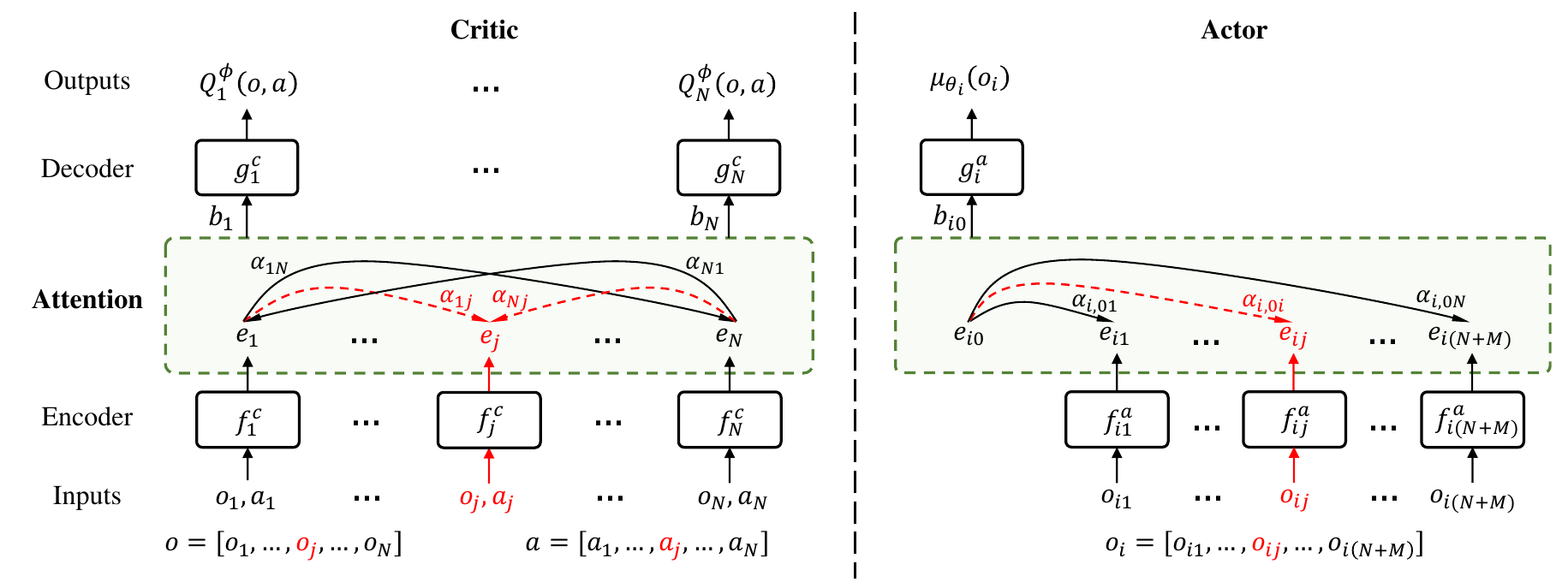}
	\caption{An illustration of the main components of our method. Actor of agent outputs action $a_{i}$ under the continuous policy $\mu_{\theta_i}$ and critic $i$ outputs $Q_i^{\phi}(o,a)$.
	Information related to the faulty agent $j$ is marked in red, and the red dashed lines represent the special attention weight for the embedding of the faulty agent.}
	\label{AACFT-figure}
	\end{center}
\end{figure*}

\section{Methods}
\label{Methods}
Our method builds upon the widely used MADDPG algorithm and can be described in three integral parts. 

First, we detail the configuration of inputs for the critic and actor to represent fault states and facilitate the use of attention mechanisms. In brief, the critic takes into account local observations and actions of all agents as input, and the actor only receives the current agent's local observations as input. Detailed descriptions of input settings are provided in \cref{Input Setting}.

Second, we design the AACFT model by incorporating attention mechanisms into the critic and actor networks, enabling it to automatically tackle the special information resulting from an unexpected agent fault, as depicted in Fig. \ref{AACFT-figure}. It is worth noting that, in such a case, the information associated with the faulty agent affects the critic and actor differently. 
For the critic, the learning algorithm needs to prioritize shifting attention away from the observations of the faulty agent and focus on other relevant information within the input. 
For the actor, the learning algorithm should be designed to make flexible decisions regarding the importance of the faulty agent at different stages of task execution. 
The attention mechanisms in the critic and actor play different roles when addressing fault issues. Therefore, it is impertive to discuss fault tolerance in the critic and actor separately, which we address in \cref{critic,actor}.

Finally, in \cref{PER}, we extend Prioritized Experience Replay (PER) to AACFT to enhance the training efficiency of both the critic and the actor, and provide the complete algorithm.

\subsection{Input Configuration}
\label{Input Setting}
Similar to the basic MADDPG algorithm, in our method, the critic of agent $i$ takes in all agents' observations $o=[o_1,...,o_N]$ and actions $a=[a_1,...,a_N]$ and outputs Q-value $Q_i^{\phi}(o,a)$. The actor of agent $i$ takes in its observation $o_i$ and outputs action $a_i$ under the continuous policy $\mu_{\theta_i}$. The Boolean variable $\mathcal{F}_{i}$ represents the fault status of agent $i$, where 1 indicates normal and 0 indicates fault.

In a multi-agent cooperation task, the observation of an agent typically encompasses the states of the following entities: the agent itself, other cooperating agents, the environment, and the task. Taking the task in Fig. \ref{introduction-figure} as an example, the state of an agent includes its position and velocity; the environment state includes the position of obstacles; the task state includes the position and velocity of the target agent. Therefore, the observation of agent $i$ can be formally described as: 
\begin{equation}
	\begin{split}
	o_i &=[o_{i1},...,o_{i(N+M)}],\\
	o_{ij} &=
		\begin{cases}
			s_i, & j=i,\\
			[Y_i(s_j),c_{ij}], & j\in\{1,...,N\}\backslash i,\\
			Y_i(s_j), & j\in\{N+1,...,N+M\},
		\end{cases}
	\end{split}
\end{equation}
where $Y_i(s_j)$ represents the agent $i$'s partial observation regarding agent $j$, and $c_{ij}$ represents the communication state from agent $i$ to agent $j$. Specifically, $c_{ij}$ is a one-hot vector, where each bit represents a different communication state. For example, in this paper, there are 3 connection states, i.e., ``normal'', ``out of communication range'', and ``fault''. Notablely, $o_{ij}$ with $j\in\{N+1,...,N+M\}$ denotes agent $i$'s partial observation of the environment or the task. 

Importantly, as shown in Fig. \ref{AACFT-figure}, the observation $o_j$ and action $a_j$ of agent $j$ become invalid if there exists an unexpected fault on agent $j$, and should be specially configured when inputting them into the critic. In addition, the normal agent $i$'s observation $o_{ij}$ regarding the faulty agent $j$ should also be specially configured when inputting it into the actor. These operations aid the algorithm in identifying special information within the input, and more details can be found in the following subsections.

\begin{algorithm*}[h]
	\caption{AACFT with priority}
	\label{Algorithm}
    \renewcommand{\algorithmicrequire}{\textbf{Input:}}
    \begin{algorithmic}[1]
    \REQUIRE batchsize $k$, replay buffer $U(\mathcal{D})$, agents number $N$, exponents $\alpha$ and $\beta$, maximum training steps $S$, replay period $K$
    \FOR {$s=1$ to $S$} 
    \STATE Step the environment and store $(x,x',a,r,\mathcal{F})$ in $U(\mathcal{D})$.
    \STATE Append priority for critic queue $\mathcal{Q}_c \leftarrow [\mathcal{Q}_c, \max{\mathcal{Q}_c}]$
	\FOR {$i=1$ to $N$}
	    \STATE Append priority for actor queues $\mathcal{Q}_{a,i} \leftarrow [\mathcal{Q}_{a,i}, \max{\mathcal{Q}_{a,i}}]$ if $\mathcal{F}_{i}=1$
	\ENDFOR
		\IF {$s \equiv 0 \ mod \ K$} 
		\STATE \emph{// Update the critics}
		\STATE Sample $k$ transitions from $U(\mathcal{D})$ to form a batch $u(\mathcal{D})$. Transition $j \sim P(j)={p_j^{\alpha}}/{\sum\nolimits_{k}p_k^{\alpha}}$
		\STATE Compute importance-sampling weights for $k$ transitions in $u(\mathcal{D})$. $\omega_j={(N\cdot P(j))^{-\beta}}/{\max_i{\omega_i}}$
		\STATE Update the critics by loss $\mathcal{L}(\phi)=\mathbb{E}_{(o,a,r,o',\omega) \sim u(\mathcal{D})} [\, \omega \cdot \sum\nolimits_{i=1}^N {\mathcal{F}_{i} \cdot (Q_i^{\phi}(o,a)-y_i)^2}]$
		\STATE Update the priorty of $k$ transitions in $\mathcal{Q}_c$
		\STATE \emph{// Update the actors}
		\FOR {$i=1$ to $N$}
			\STATE Sample $k$ transitions from $U(\mathcal{D})$ to form a batch $u(\mathcal{D})$. Transition $j \sim P(j)={p_j^{\alpha}}/{\sum\nolimits_{k}p_k^{\alpha}}$
			\STATE Compute importance-sampling weights for $k$ transitions in $u(\mathcal{D})$. $\omega_j={(N\cdot P(j))^{-\beta}}/{\max_i{\omega_i}}$
			\STATE Update the actor of agent $i$ by loss $\mathcal{L}(\theta_i)=-\mathbb{E}_{(o,a,\omega) \sim u(\mathcal{D})}
			[\, \omega \cdot Q_i^\phi(o,(a_i,a_{\backslash i}))]|_{a_i=\mu_{\theta_i}(o_i)}$
			\STATE Update the priorty of $k$ transitions in $\mathcal{Q}_{a,i}$
		\ENDFOR
		\ENDIF
    \ENDFOR
    \end{algorithmic}
\end{algorithm*}

\subsection{Fault Tolerance in Critic}
\label{critic}
Unexpected agent faults can directly impact the input of the critic, as the faulty agent $j$ becomes unable to observe or move, rendering the original $o_j$ and $a_j$ values invalid. To enable the critic to distinguish whether an agent fails or not, we set $o_j=z\cdot\mathbf{1},a_j= \mathbf{0}$ when $\mathcal{F}_{j}=0$, i.e., agent $j$ fails. Each bit of $o_j$ is a special flag $z$, the absolute value of which should be much larger than the normal values to make the anomalies more prominent in the input. While this setting benefits the critic in detecting the occurrence of faults, the excessive value of special flag $z$ significantly disrupts critic network training, especially when the faults happen randomly.

To tackle this issue, we introduce an attention module into the structure of the critic. This module helps in diverting attention from the observations of the faulty agent and instead focuses on other pertinent information within the input. As shown in Fig. \ref{AACFT-figure}, the encoder $f^c_{i}$ transforms the observation and action of agent $i$ into its embedding $e_i=f^c_{i}(o_i,a_i)$. The attention module calculates the attention weight $\alpha_{ij}$ to the embeddings $e_j$ according to Equ. (\ref{att1}), and the attention embedding $b_i$ used for evaluating the Q-value of agent $i$ according to Equ. (\ref{att2}). Then, the decoder $g^c_{i}$ extracts agent $i$'s Q-value $Q_i^{\phi}(o,a)$ from $b_i$: 
\begin{equation}
	Q_i^{\phi}(o,a)=g^c_{i}(b_i)=g^c_{i}(\textstyle\sum\nolimits_{j=1}^{N}\alpha_{ij} W_v e_{j}),
\end{equation}
where $\phi$ denotes the parameter set of critics, including the individual parameters of all agents' encoders and decoders, and the public parameters of $W_q, W_k, W_v$ in Equ. \ref{att1}.

All critics are updated together to minimize a joint regression loss function $\mathcal{L}(\phi)$ due to parameter sharing: 
\begin{equation}
	\begin{split}
	\mathcal{L}(\phi)=\mathbb{E}_{U(\mathcal{D})} [\sum\nolimits_{i=1}^N {\mathcal{F}_{i} \cdot (Q_i^{\phi}(o,a)-y_i)^2}],\\
	y_i=r_i+\gamma Q_i^{\phi'}(o',a'_1,...,a'_N)|_{a'_j=\mu_{\theta'_j}(o_j)},
	\end{split}
	\label{loss_critic}
\end{equation}
where ${\phi'}$ and ${\theta'}$ denote the delayed parameters of the target critics and target policies, respectively. The actor of the faulty agent is disabled, so that its critic which guides the actor's update also becomes meaningless. Therefore, the faulty agents are excluded using $\mathcal{F}_{i}$ when calculating the loss.

\subsection{Fault Tolerance in Actors}
\label{actor}
Unexpected agent faults can also have a direct impact on the input of the actor, as the faulty agent $j$'s state could affect the observations of the remaining normal agents, i.e., $o_{ij}$, where $i\in\{1,...,N\}\backslash j$. 
To enable the actor to distinguish the agent fault and selectively use the state information of the faulty agent, we set the value of $o_{ij}$ according to the significance of the faulty agent $j$ at different stages of task execution. 
Specifically, if the state of faulty agent $j$ still makes sense to decisions of other agents after it fails, then $o_{ij}=Y_{i}(s_{j,t_0})$, where $t_0$ is the time step fault occurs. 
Otherwise, similar to the critic network, we set $o_{ij}=z\cdot\mathbf{1}$ to make the anomalies more prominent and easier to identify. 

According to the above, the sudden occurrence of agent faults places high demands on the performance of the actor network, 
requiring it to possess the ability to dynamically determine whether to prioritize attention to the faulty agent at different stages of task execution. To reach this goal, 
in addition to the special design of the input state representation, we innovatively introduce an attention module in the actor to flexibly incorporate information about the faulty agent. 
As shown in Fig. \ref{AACFT-figure}, different from the structure of the critic network, an actor aims to output one action for the whole observation space. Therefore, referring to models with attention for classification \cite{VIT}, we add a token $e_{i0}$ for agent $i$ to help output the action. 
On such a basis, the encoder $f^a_{ij}$ transforms $o_{ij}$ into its embedding $e_{ij}=f^a_{ij}(o_{ij})$. 
The attention module calculates the attention weight $\alpha_{i,0j}$ of the token $e_{i0}$ to $e_{ij}$ and its attention embedding $b_{i0}$. Then, the decoder $g^a_{i}$ outputs agent $i$'s action $a_i$ from $b_{i0}$:
\begin{equation}
	\mu_{\theta_i}(o_i)=g^a_{i}(b_{i0})=g^a_{i}(\textstyle\sum\nolimits_{j=1}^{N+M}\alpha_{i,0j} W_v^i e_{ij}),
\end{equation}
where $\theta_i$ is the parameter set of agent $i$'s actor, which includes the parameters of all encoders and the decoder, and the public parameters of $W_q^i,W_k^i,W_v^i$.

To update agent $i$'s actor, $a_i$ is removed from the input of $Q_i^\phi$ and replaced with $\mu_{\theta_i}(o_i)$, while other agents' actions $a_{\backslash i}$ are fixed. Each actor is updated to minimize its loss, i.e., the opposite of the value of the new action combination:
\begin{equation}
	\mathcal{L}(\theta_i)=-\mathbb{E}_{U(\mathcal{D})}
	[\mathcal{F}_{i}\cdot Q_i^\phi(o,(a_i,a_{\backslash i}))]|_{a_i=\mu_{\theta_i}(o_i)}.
	\label{loss_actor}
\end{equation}
Similar to the situation with the critic, due to the disablement of the faulty agent's actor, the corresponding transitions are excluded.

\begin{figure*}[tbp]
	\begin{center}
	\centering
	\includegraphics[width=0.8\linewidth]{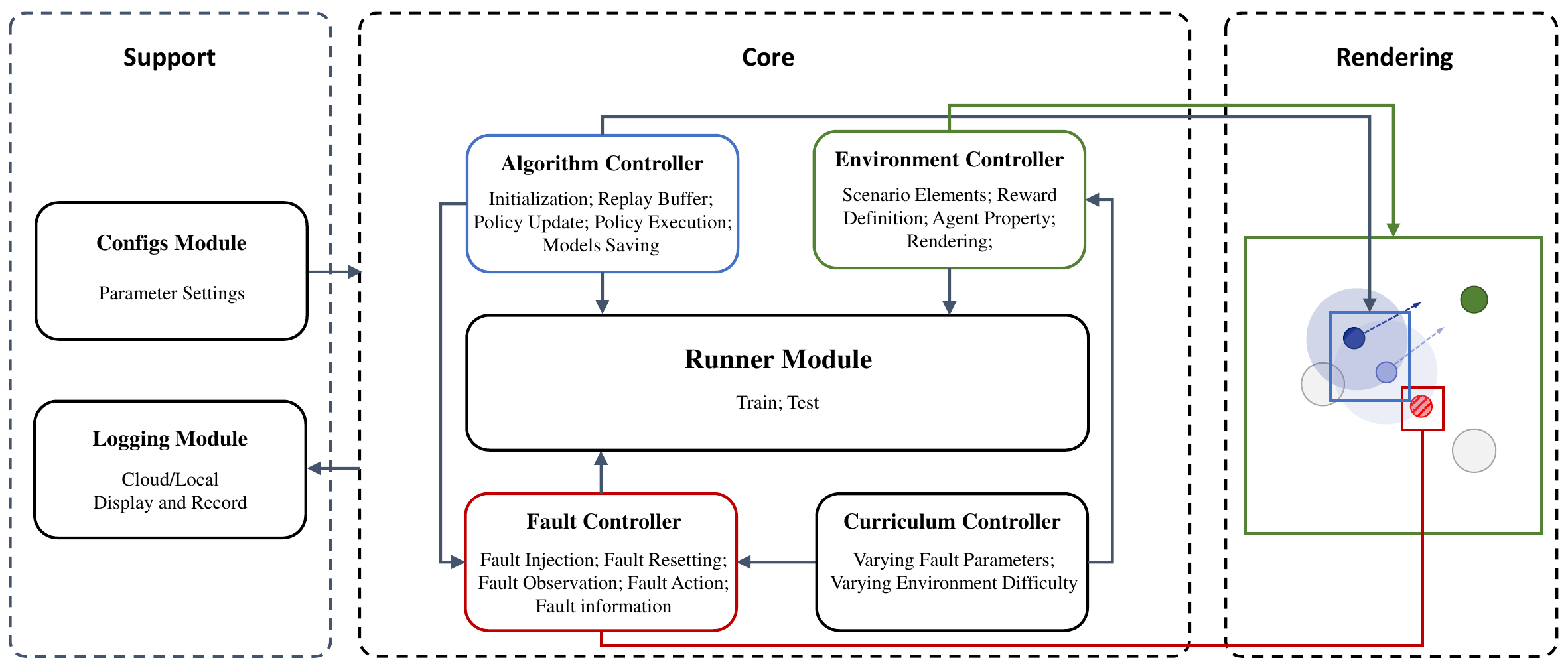}
	\caption{The overall framework of FTMAL.}
	\label{FTMAL}
	\end{center}
\end{figure*}

\subsection{Prioritized Experience Replay for AACFT}
\label{PER}
Imbalanced transitions stored in the replay buffer are reweighted to improve training efficiency. As mentioned in Section \ref{critic} and Section \ref{actor}, $N+1$ modules conduct updates. To apply PER to AACFT, PER is employed across all modules. A transition exhibits different importance for each module, so different priorities should be assigned. 
In AACFT, the critics are updated simultaneously due to parameter sharing, so a single priority queue is set up for them, named as $\mathcal{Q}_c$. On the other hand, the actors of different agents are updated separately, so each actor is assigned its own priority queue, named as $\{\mathcal{Q}_{a,1},...,\mathcal{Q}_{a,N}\}$.

When the transition $(x,x',a,r,\mathcal{F})$ is stored, it is assigned a corresponding priority in each queue, with the value being the highest priority currently in the queue. It is important to note that if the transition involves faulty agent $i$, the priority queue $\mathcal{Q}_{a,i}$ does not add a new priority for this transition, to avoid affecting the priority ranking and prevent the transitions from being sampled when updating agent $i$'s actor.

For $\mathcal{Q}\in\{\mathcal{Q}_c, \mathcal{Q}_{a,1},...,\mathcal{Q}_{a,N}\}$, denote $p_i$ as the priority of  transition $i$. $p_i = 1/\text{rank}(i)$ is rank-based priority of transition $i$. The rank of a transition is sorted according to its loss $\mathcal{L}(\phi)$ or $\mathcal{L}(\theta)$. So the priority is monotonic with the loss.
The probability of sampling transition $i$ is 
\begin{equation}
	P(i)=\frac{p^{\alpha}_i}{\sum\nolimits_{k}p^{\alpha}_k}.
	\label{sample_probability}
\end{equation}
The exponent $\alpha$ determines how much prioritization is used, with $\alpha=0$ corresponding to the uniform case.

To ensure the unbiased nature of the updates at the end of training, 
loss is weighted by importance-sampling weights
\begin{equation}
	\omega_j=\frac{(N\cdot P(j))^{-\beta}}{\text{max}_i{\omega_i}},
	\label{sample_weight}
\end{equation}
where $\beta$ is linearly annealed from its initial value to 1.

The algorithm is described in Algorithm \ref{Algorithm}.

\section{A New Platform for Fault-Tolerant MARL}
\label{platform}
Current reinforcement learning platforms integrate multiple algorithms and environments \cite{rllib17, rlpyt19}, facilitating the study of different algorithms' performance across various environments. However, they present challenges when adapting to fault-tolerant MARL algorithms. Faults can affect multiple aspects of experiments, creating strong dependencies between different modules in the code. For instance, in the main program, when the environment advances to the next timestep, it must decide whether to inject faults. After faults are introduced, the scenario scripts need to simulate abnormal observations or actions caused by the faults, while the algorithm scripts may need to filter and process data from faulty agents. Existing platforms face low efficiency in training and testing fault-tolerant MARL models due to these complexities.

We design and open-source a new platform for fault-tolerant MARL, with its framework shown in Fig. \ref{FTMAL}. The platform consists of a core structure and supporting modules, with a rendered environment diagram provided to clarify the functionality of each controller. To overcome the aforementioned limitations, our core approach is to centralize the management of faults in a dedicated controller, referred to as the Fault Controller. This design ensures that fault modifications do not interfere with the main components of the code. 

The roles of each component in the platform are as follows. The core structure revolves around the runner module, which handles the fundamental processes of MARL, including agent-environment interactions and the collection of transitions through the replay buffer. 

Multiple controllers manage different aspects of the experiment. The Algorithm Controller is responsible for initializing, updating, and executing agent policies, as well as handling fault-related tasks such as processing observations. The Environment Controller defines the environment, scenarios, and agents, and also handles the rendering of the environment. The Fault Controller provides functionalities for fault injection, modification of faulty agents' observations and actions, fault resetting, and the collection of fault-related data. Finally, the Curriculum Controller offers interfaces for curriculum learning, enabling the setup of varying faults throughout the training process.

The Configs Module and Logging Module provide essential support for the main experimental framework. The Configs Module supplies the relevant parameters for algorithms, environments, faults, and training or testing settings. It supports separate configuration files for algorithms and environments, making it easier to conduct comparative experiments. The Logging Module records data from the algorithm, environment, and fault controllers. It supports both WandB \cite{wandb} integration and local options for printing and saving training logs.

Users have the flexibility to customize fault types and corresponding parameters, such as fault timing, target, and probability. Fault types include restricting the actions of faulty agents or adding noise to their observations. Due to the decoupling of faults from the environment and algorithm code, users can inject the same type or multiple types of faults across different scenarios, allowing them to investigate the performance of various algorithms in handling different fault conditions.
\begin{figure*}[h]
	\begin{center}
	\centering
	\includegraphics[width=0.95\linewidth]{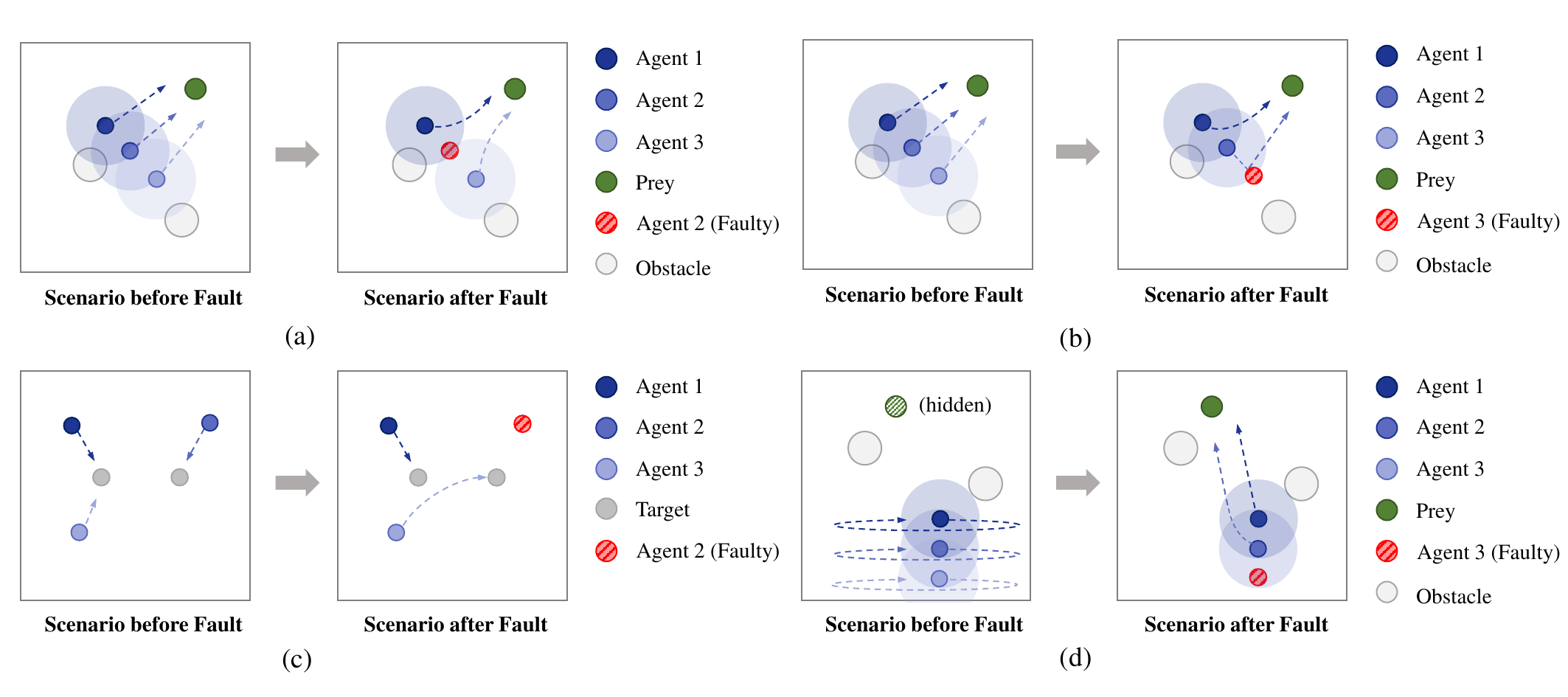}
	\caption{Schematic diagram of the scenarios before and after faults. (a) Abandonment scenario and recovery scenario if agent 2 fails. (b) Recovery scenario if agent 3 fails. (c) Navigation scenario. (d) Patrol scenario. }
	\label{scenarios}
	\end{center}
\end{figure*}

\section{Experiments}
\label{Experiments}
\subsection{Setup}
\subsubsection{Experiments Overview}
This section evaluates and validates our proposed methods from various perspectives. In Subsection \ref{subsec:1}, we assess the performance of a multi-agent system trained without considering potential faults when a fault occurs, demonstrating the necessity of enhancing fault-tolerant capabilities of RL algorithms. In Subsection \ref{subsec:2}, we compare the AACFT proposed in this paper with three baseline methods, illustrating that AACFT can improve the fault tolerance of MARL. 
In Subsection \ref{subsec:3}, we visualize an episode of a scenario to show how the trained agents execute actions to mitigate the adverse effects of faults. In Subsection \ref{subsec:4}, we visualize the attention distribution to validate the role of attention in AACFT. We analyze the changes in attention before and after faults to understand its impact, and we conduct ablation studies by removing attention in the critic and actor modules separately.
In Subsection \ref{subsec:5}, to verify the effectiveness of PER, we conduct comparative experiments and analyze the changes in the number of different types of transitions sampled in the training batches. Finally, in Subsection \ref{subsec:6}, we test the performance of the trained model at different fault timesteps to demonstrate AACFT's adaptability to temporal variations.

\subsubsection{Scenarios}
To comprehensively validate the experimental objectives presented in the previous subsection, we design four scenarios modified from Multi-Agent Particle Environment (MPE) \cite{MADDPG}. Schematic diagrams are shown in Fig. \ref{scenarios}.

\textbf{Abandonment Scenario }
The abandonment scenario is a fault-tolerant variation of the basic predator-prey setup. At the start of each episode, we randomly generate two obstacles, three cooperative agents (``predators'') to be trained, and one target agent (``prey'') to be pursued. In this scenario, faults may disrupt the communication structure among agents, potentially reducing the number of functioning agents in the system. The remaining agents must maintain communication and continue pursuing the target, while disregarding and abandoning any faulty agents. 

\textbf{Recovery Scenario }
In addition to potential communication breakdowns, faults may also introduce temporary tasks for the agents. To validate our method, we design a scenario called the Recovery Scenario. This scenario mirrors the abandonment scenario when agent 2 fails, but with a key difference: agent 3 holds valuable resources that must be recovered by agent 2 in the event of a fault. In our experiments, a successful recovery is defined as agent 2 touching the faulty agent. We assume agent 2 can recover the resources from the faulty agent 3 and continue the pursuit. The distinct roles of agents 2 and 3 also test the system's ability to handle asymmetric faults.

\textbf{Navigation Scenario }
In the navigation scenario, agents are required to dynamically reschedule between multiple task objectives, testing their ability to allocate resources or targets effectively. When no faults occur, three agents are tasked with covering two landmarks. The challenge in this scenario arises when a fault unexpectedly occurs, requiring the remaining agents to reassess their target landmarks and adjust based on the current states of the agents and the landmarks. 

\textbf{Patrol Scenario }
In the patrol scenario, faults introduce entirely new targets and tasks for the agents, challenging the balance of training for tasks before and after a fault occurs. When no faults are present, three agents are tasked with efficiently patrolling the bottom half of the map. However, an enemy agent randomly attacks one of the patrolling agents, causing it to fail. The remaining two agents then detect the location of the faulty agent and initiate a pursuit.

The reward $r$ of each agent is composed of several parts:
\begin{equation}
	r=\underbrace{r_{\text{com}}+r_{\text{out}}+r_{\text{col}}+r_{\text{dis}}}_{r_{\text{indi}}}+\underbrace{r_{\text{goal}}}_{r_{\text{team}}},
\end{equation}
where $r_{\text{com}},r_{\text{out}},r_{\text{col}},r_{\text{dis}}$ 
are rewards encouraging the agent to maintain the communication structure, stay within the boundary, avoid collisions with teammates, and approach the target, respectively.
These four rewards are individual rewards and are received by each agent independently, 
whereas $r_{\text{goal}}$ is a team reward, granted to all agents when any one of them completes the task.

\begin{figure}[htbp]
	\centering
	\hspace{-0.3cm}
	\begin{minipage}[\label{fig:a}]{0.42\linewidth}
		\centering
		\includegraphics[width=\linewidth]{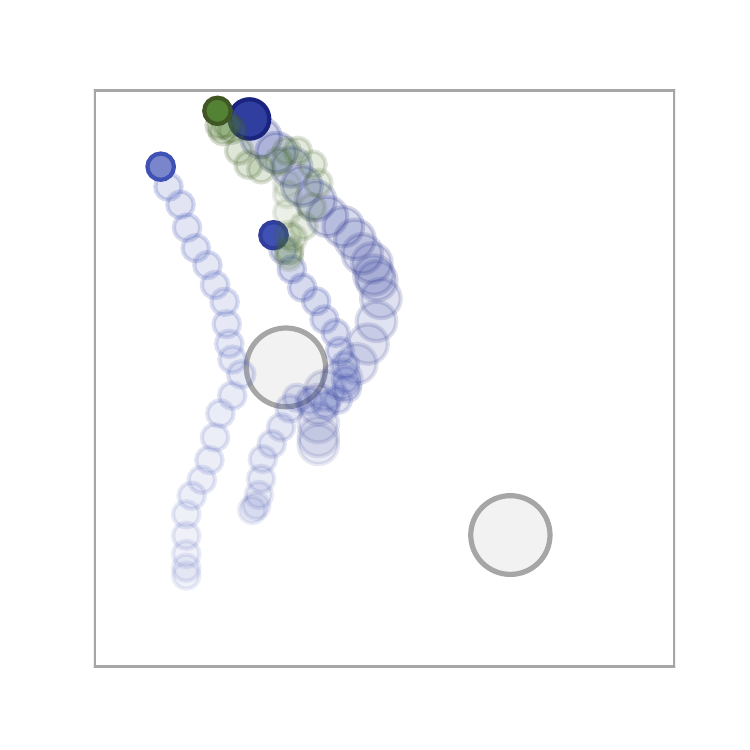}
	\end{minipage}
	\hspace{-0.6cm}
	\begin{minipage}[\label{fig:b}]{0.42\linewidth}
		\centering
		\includegraphics[width=\linewidth]{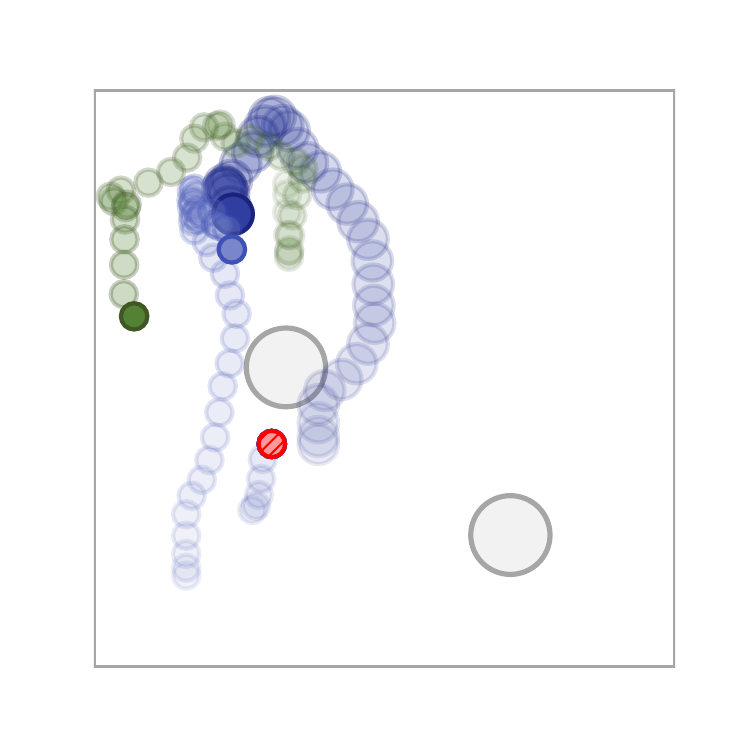}
	\end{minipage}
	\hspace{-0.3cm}
    \begin{minipage}{0.21\linewidth}
		\centering
		\includegraphics[width=\linewidth]{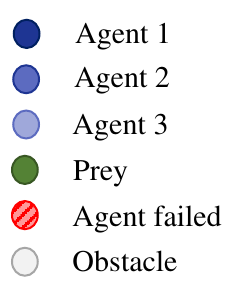}
	\end{minipage}
	\vspace{-0.3cm}
	\caption{Testing of the basic MADDPG without considering faults in the training process.
			(Left) No fault. (Right) Agent 2 fails.}
	\label{nofault-figure}
\end{figure}

\begin{figure*}[htbp]
	\centering
	\subfloat[\label{fig:a}]{
		\includegraphics[width=0.35\linewidth]{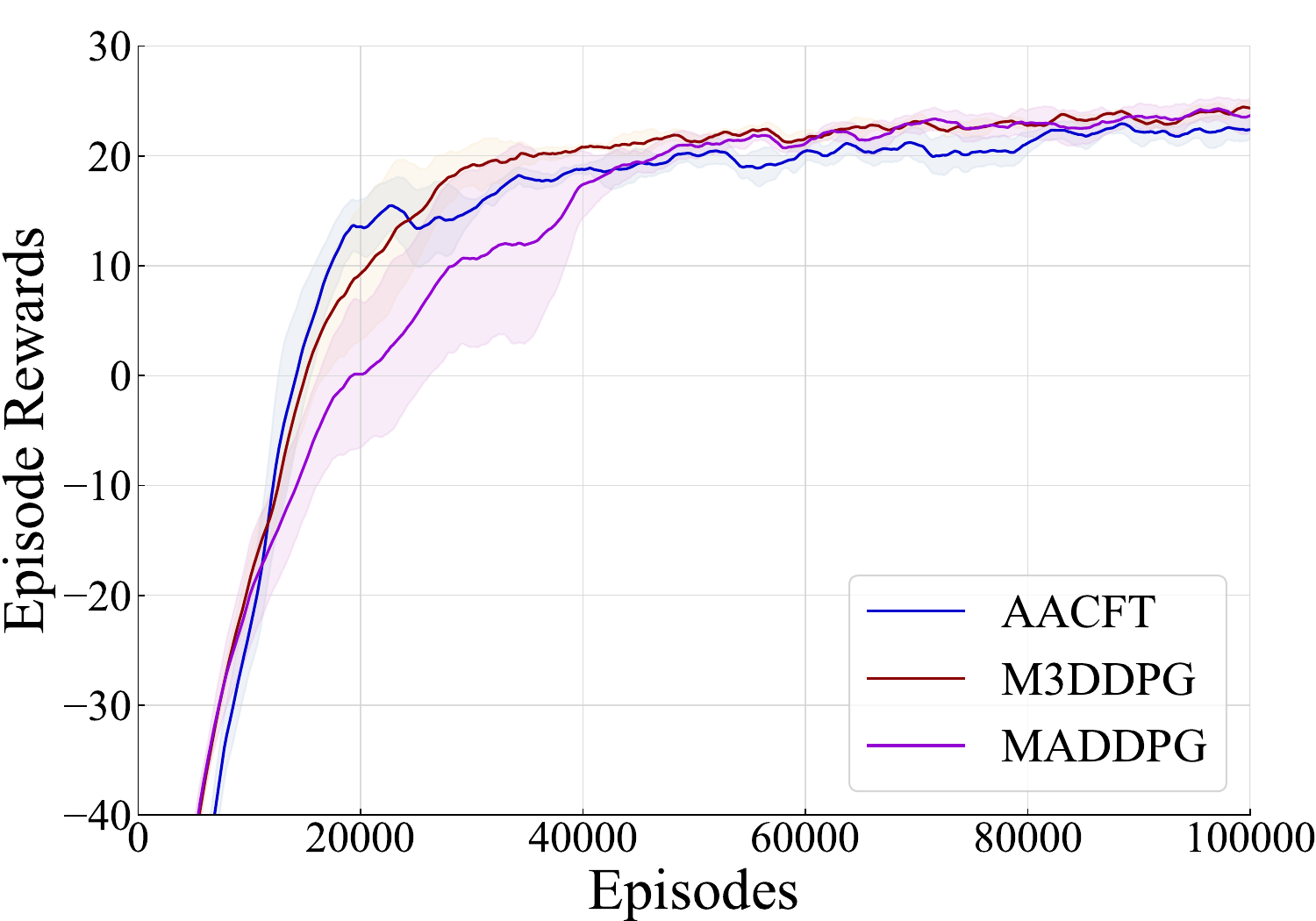}}
		\hspace{0.6cm}
	\subfloat[\label{fig:b}]{
		\includegraphics[width=0.35\linewidth]{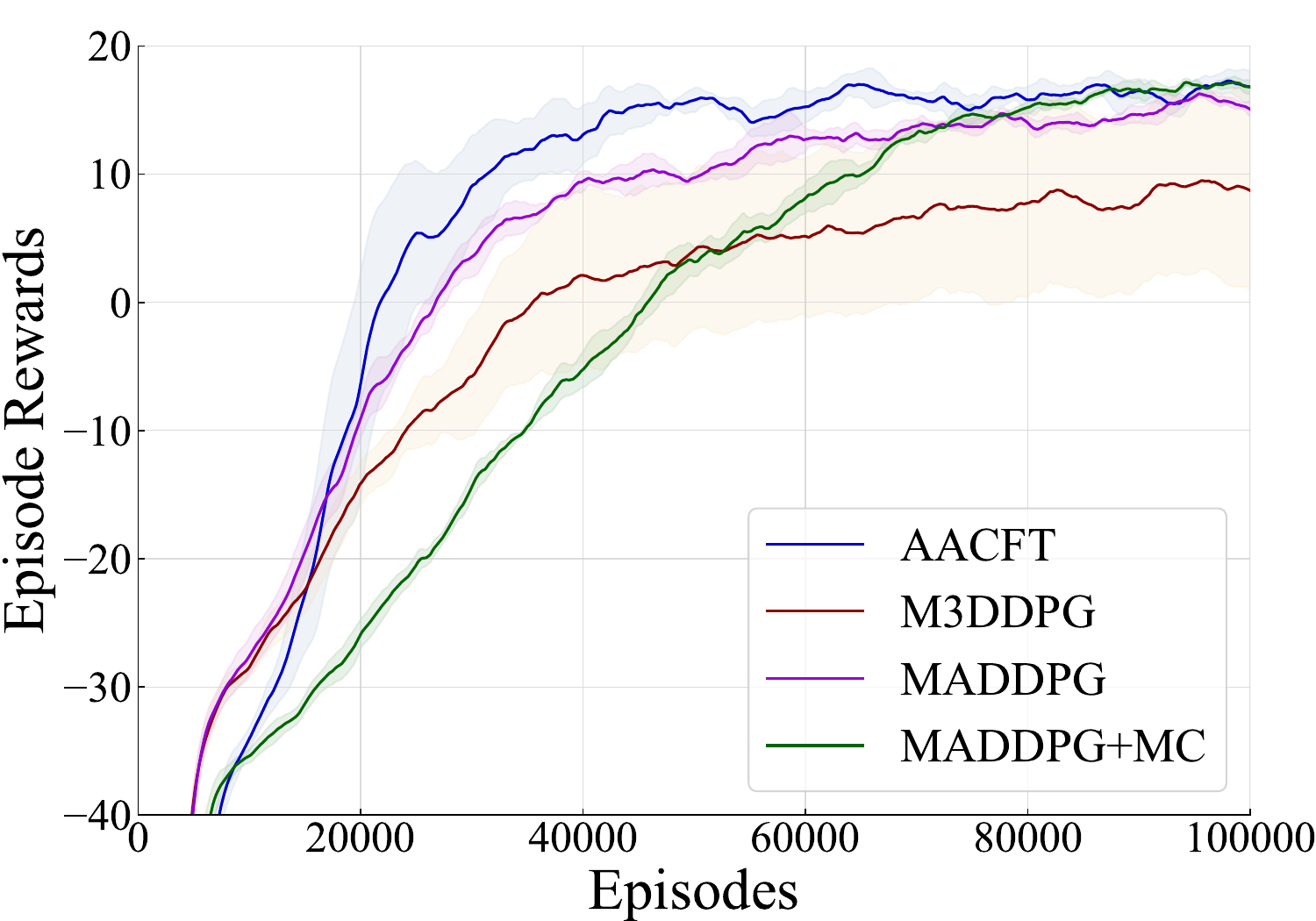}}\\
	\subfloat[\label{fig:c}]{
		\includegraphics[width=0.35\linewidth]{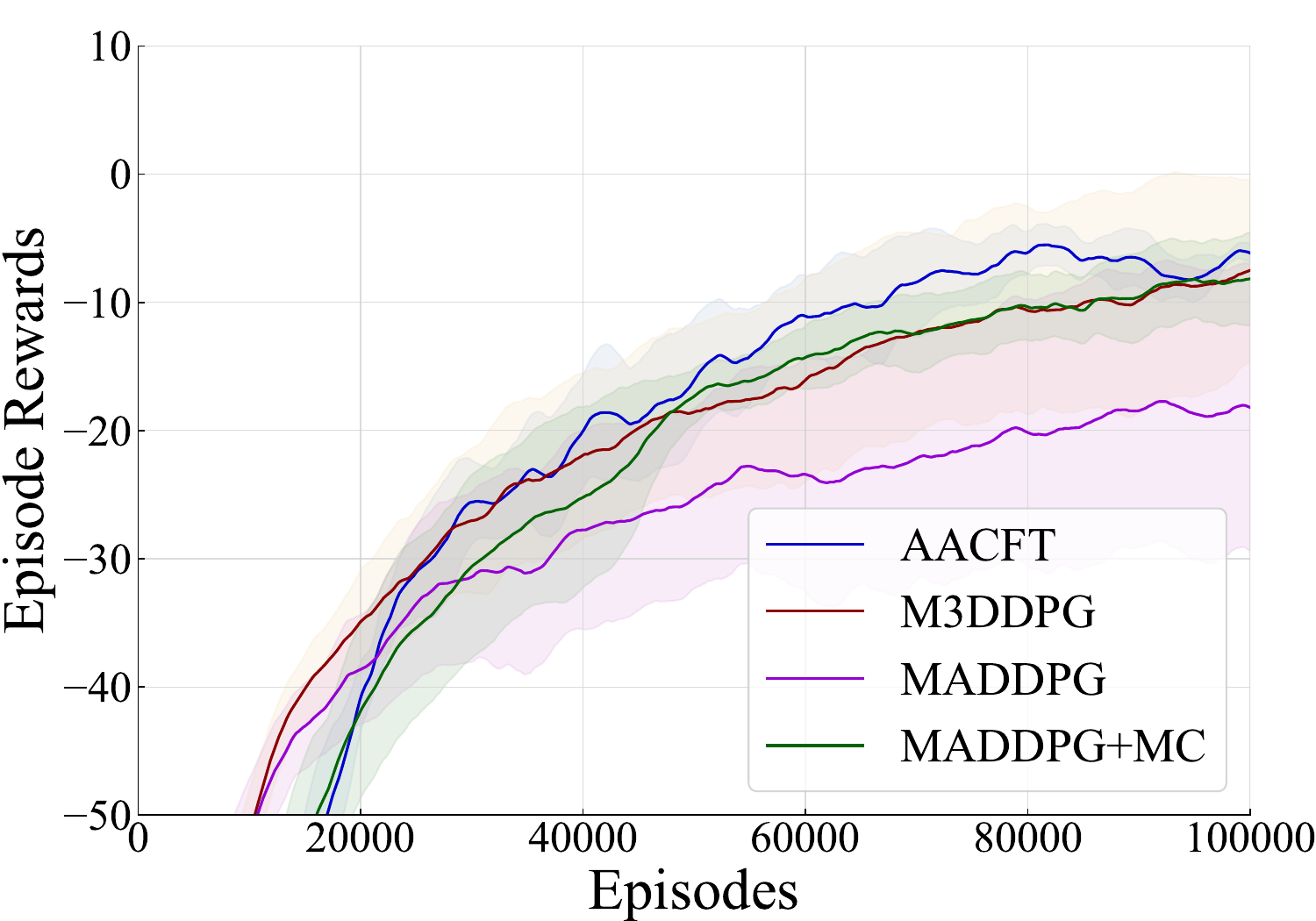}}
		\hspace{0.6cm}
	\subfloat[\label{fig:d}]{
		\includegraphics[width=0.35\linewidth]{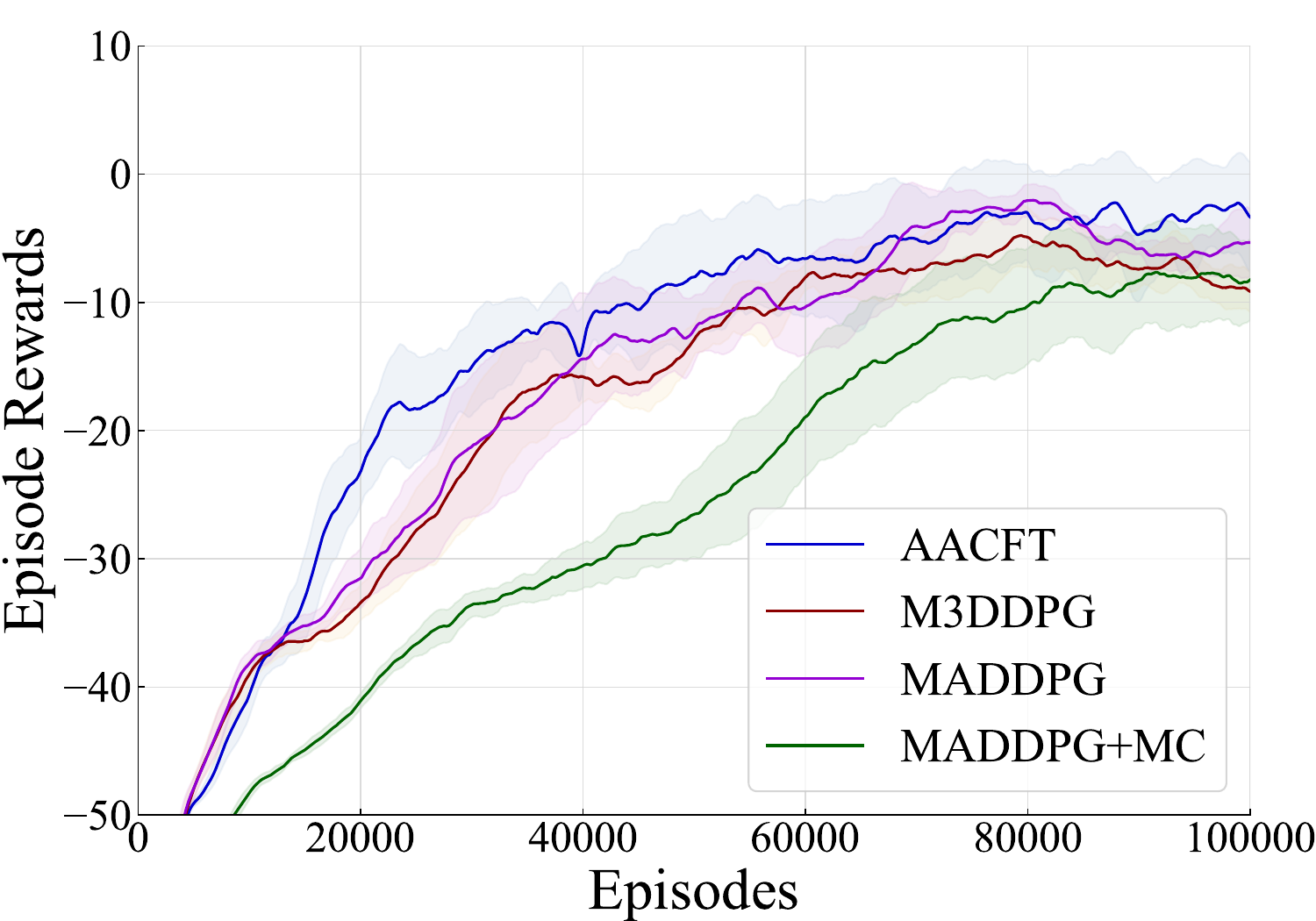}}
		\caption{Reward curves from 4 training sessions for each method in 4 scenarios.
		(a) No-fault predator-prey scenario;
		(b) Abandonment scenario;
		(c) Recovery scenario;
		(d) Navigation scenario.
	}
	\label{result-curve}
\end{figure*}
\begin{figure*}[htbp]
	\centering
	\subfloat[\label{fig:a}]{
		\includegraphics[width=0.33\linewidth]{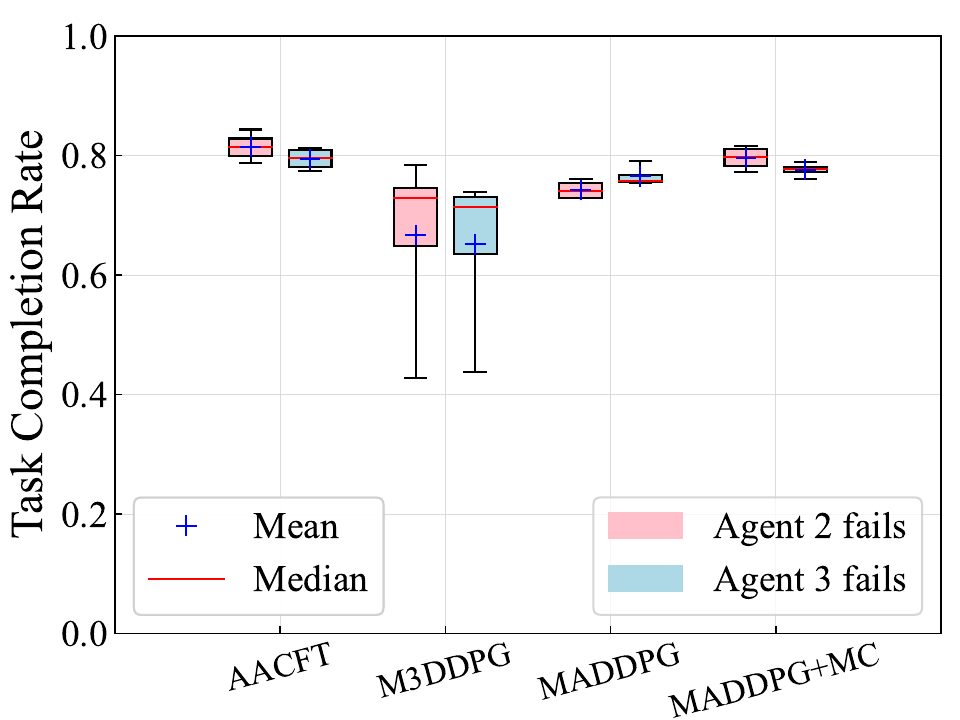}}
	\subfloat[\label{fig:b}]{
		\includegraphics[width=0.33\linewidth]{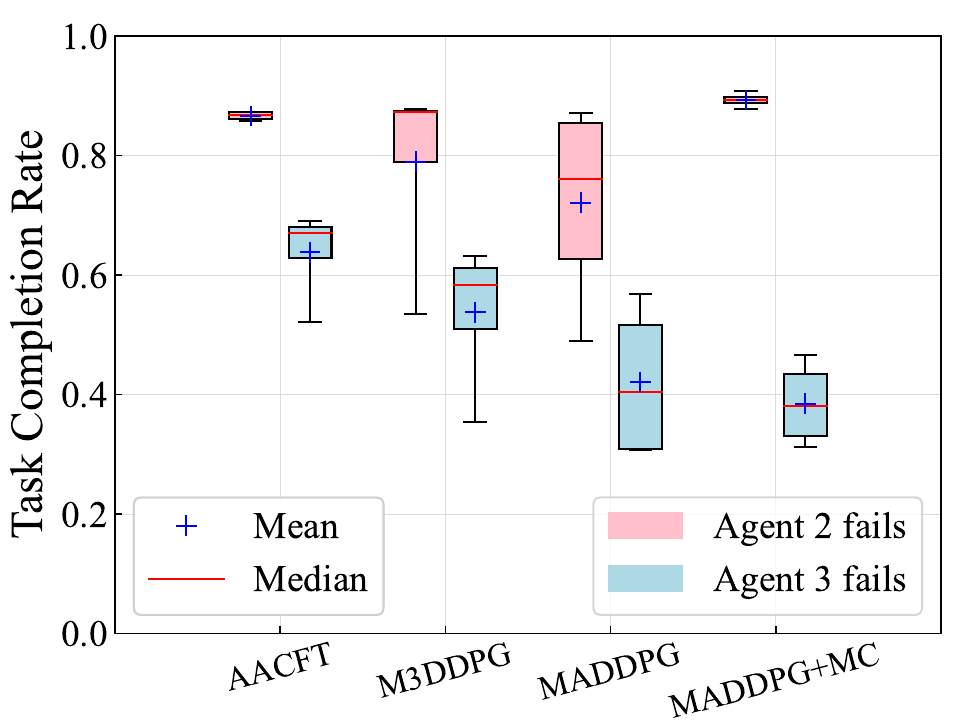}}
	\subfloat[\label{fig:c}]{
		\includegraphics[width=0.33\linewidth]{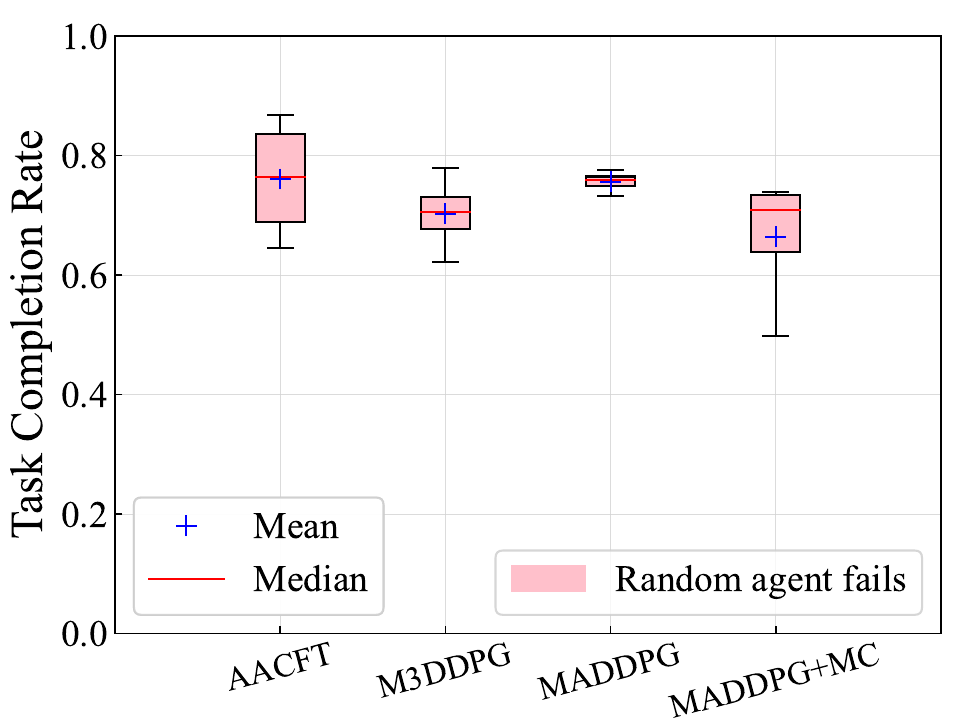}}
		\caption{Boxplots of evaluations results from 4 training sessions for each method in 3 scenarios. 
		(a) Abandonment scenario;
		(b) Recovery scenario;
		(c) Navigation scenario.
	}
	\label{result-boxplot}
\end{figure*}

\subsection{Results and Analysis}
\subsubsection{Necessarity}\label{subsec:1}
We use vanilla MADDPG to train the multi-agent system in the basic predator-prey scenario without considering potential faults.
Compared to the test in a no-fault environment, the task completion rate drops from 0.872 to 0.382 in the test where agent 2 fails at time step 5. This demonstrates the necessity of a fault-tolerant model.
Fig. \ref{nofault-figure} shows a comparison of the model's performance in environments with and without faults. 
Without faults, the system completes the task at time step 23. 
When agent 2 fails, agents 1 and 3 can still approach the prey, but they become stuck in a state of ``helplessly wandering in place'' in the upper left corner. This is because they lack the ability to identify faults and reassign tasks. As a result, the prey seizes the opportunity to escape downwards. Ultimately, the system fails to complete the task within the allotted 35 time steps.

\begin{figure*}[htbp]
	\centering
	\begin{minipage}{0.85\linewidth}
		\centering
		\includegraphics[width=\linewidth]{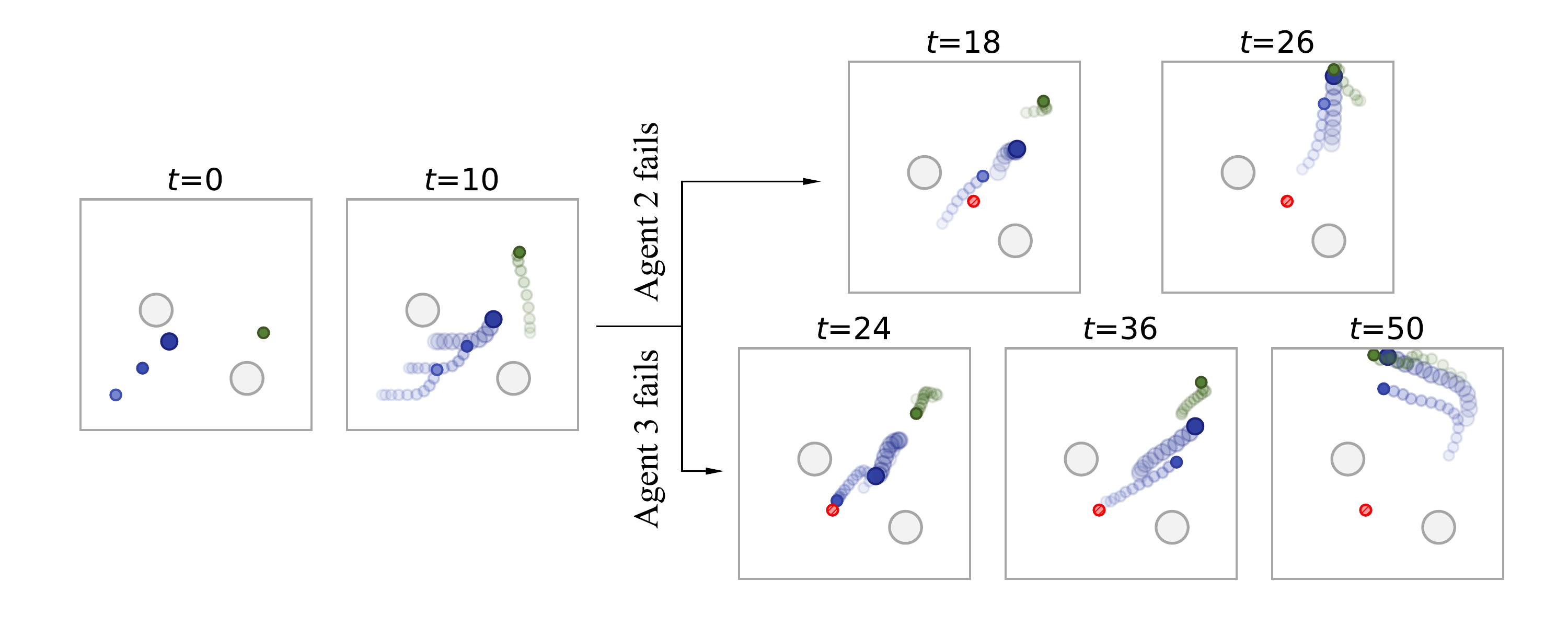}
	\end{minipage}
    \begin{minipage}{0.12\linewidth}
		\centering
		\includegraphics[width=\linewidth]{figures/legend.pdf}
	\end{minipage}
	\caption{Illustration of an episode in the recovery scenario,
			where agent 2 and agent 3 fail respectively. In the figure, a series of circles with gradually decreasing transparency represents the motion trajectory of the agent from the previous time step to the current time step.}
	\label{recovery-figure}
\end{figure*}
\begin{figure*}[h]
	\centering
		\includegraphics[width=0.95\linewidth]{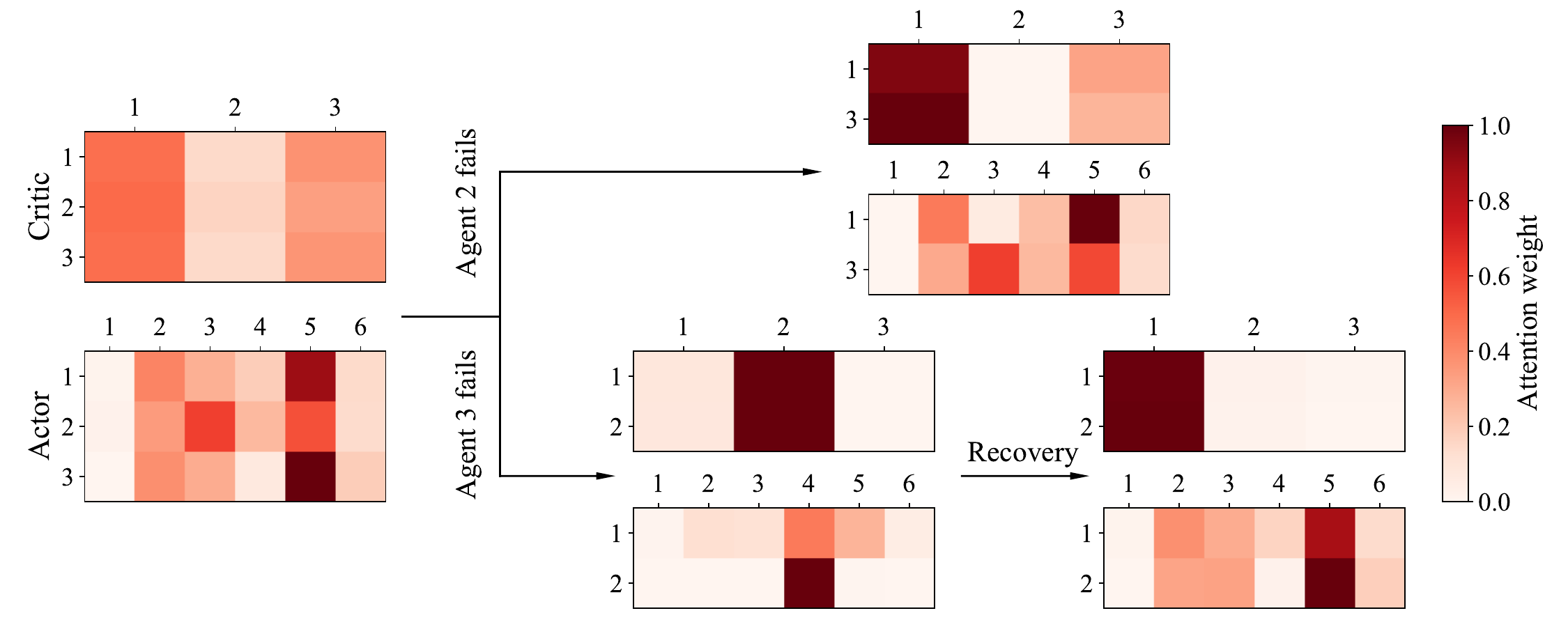}
	\caption{Attention allocation on critic and actor before and after fault in the recovery scenario. 
	The $3\times3$ or $2\times3$ matrix is the attention matrix for the critic, and the $3\times6$ or $2\times6$ matrix is the attention matrix for the actor. 
	Each row in the critic represents the attention allocation of an agent to other agents, 
	and each row in the actor represents the attention allocation of an agent to different information in its observation. 
	After the fault, the corresponding agent row is removed.}
	\label{attention-figure}
\end{figure*}

\subsubsection{Comparative Experiments}
\label{subsec:2}
Fig. \ref{result-curve} shows the training process among AACFT and three baseline methods: M3DDPG, MADDPG and MADDPG+MC (with multiple critics).
Given that the patrol scenario poses a challenge for both vanilla AACFT and the baseline methods, we conduct comparative experiments in the first three scenarios.

M3DDPG is a typical method of robust MARL. It is a MiniMax extension of MADDPG, preventing agents from learning the local optima of other agents' current strategies, thereby enabling the trained agents to generalize effectively even when other agents' strategies alter.
MADDPG is the automatic identification method shown in Fig. \ref{introduction-figure}(d) right. Each agent possesses a critic and an actor, which handle both pre-fault and post-fault scenarios. 
MADDPG+MC is the manual identification method shown in Fig. \ref{introduction-figure}(d) left. Each agent possesses three critics, one for pre-fault scenarios, two for the faults of the other two agents respectively. Accordingly, transitions are collected into corresponding replay buffers, and critics are separately trained. To reduce the number of networks and facilitate comparison, we do not establish multiple actors before and after faults.

As depicted in Fig. \ref{result-curve}, all methods achieve similar results in no-fault scenarios. However, except for our AACFT, other methods show a decline in rewards in either the abandonment or recovery scenario. We construct boxplots using task completion rate as the metric shown in Fig. \ref{result-boxplot}. 
M3DDPG underperforms compared to AACFT in all scenarios, indicating that methods aimed at increasing robustness do not directly address fault tolerance issues. MADDPG's average performance is close to AACFT only in the navigation scenario, showing that the performance of MADDPG is affected by invalid fault information. MADDPG+MC only surpasses AACFT in the recovery scenario when agent 2 fails, but performs poorly otherwise. Additionally, MADDPG+MC requires significant space resources and is overly cumbersome. Notably, when agent 3 fails in the recovery scenario, AACFT shows the most significant performance improvement, demonstrating its effectiveness in handling complex issues such as task redistribution caused by faults. Overall, in different scenarios where faults have various impacts, AACFT effectively addresses handle the issues caused by the faults.

\begin{figure*}[h]
	\centering
	\hspace{0.1cm}
	\subfloat[\label{fig:a}]{
		\includegraphics[width=0.35\linewidth]{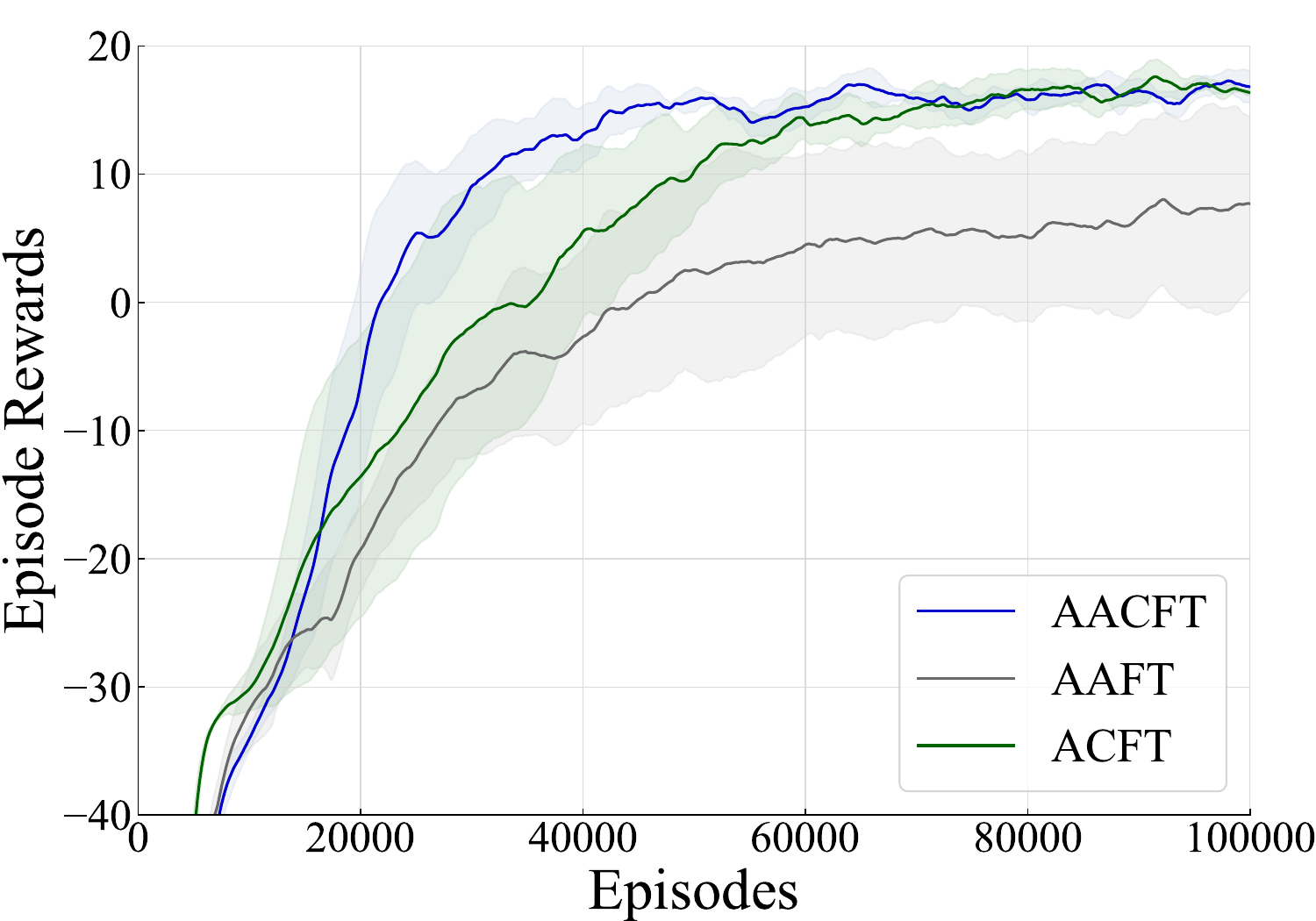}}
		\hspace{0.6cm}
	\subfloat[\label{fig:b}]{
		\includegraphics[width=0.35\linewidth]{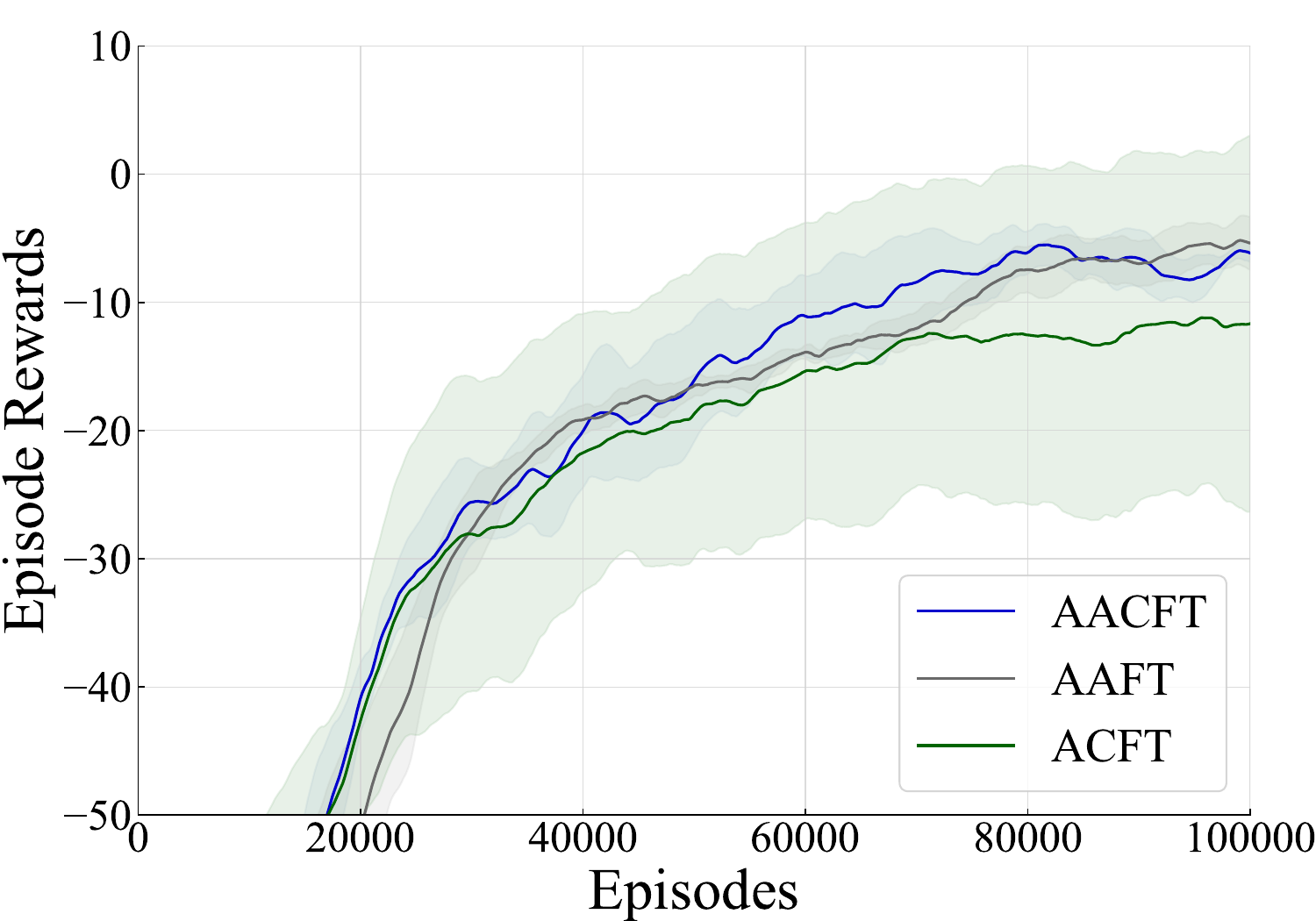}}
		\caption{Reward curves from 4 training sessions for each method in 2 scenarios.
		(a) Abandonment scenario;
		(b) Recovery scenario;
	}
	\label{result-curve-ablation}
\end{figure*}
\begin{figure*}[h]
	\centering
	\subfloat[\label{fig:a}]{
		\includegraphics[width=0.33\linewidth]{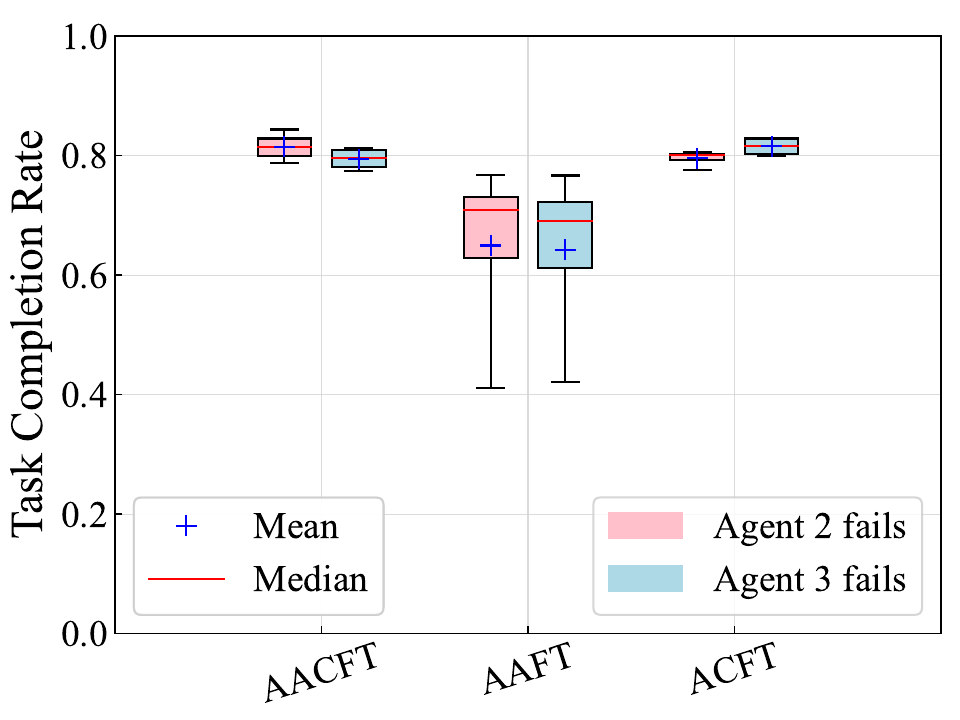}}
		\hspace{0.9cm}
	\subfloat[\label{fig:b}]{
		\includegraphics[width=0.33\linewidth]{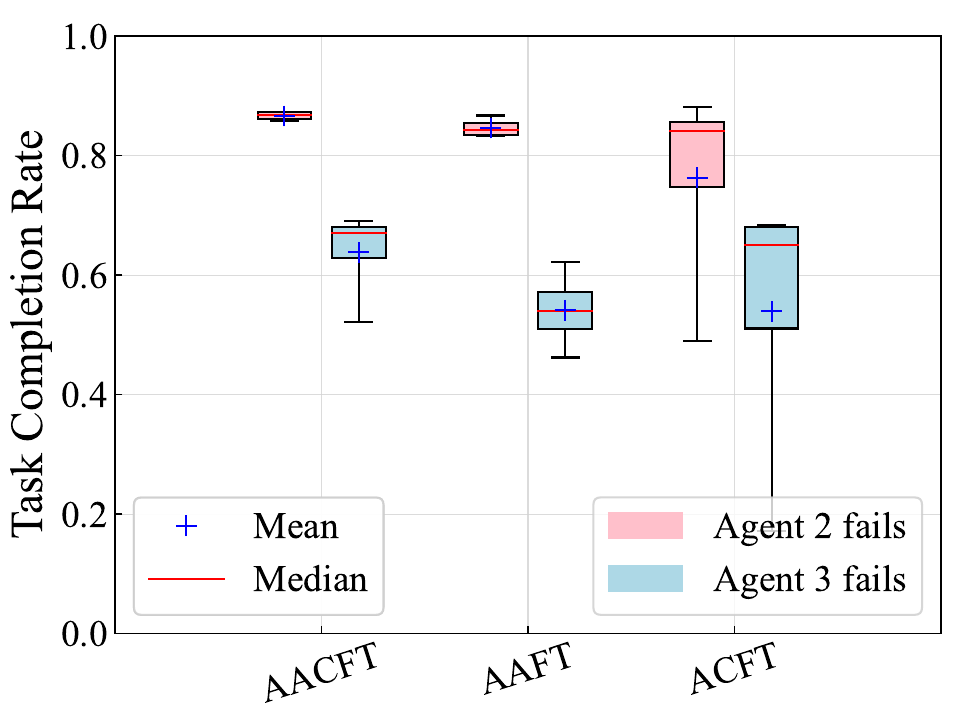}}
		\caption{Boxplots of evaluations results from 4 training sessions for each method in 2 scenarios. 
		(a) Abandonment scenario;
		(b) Recovery scenario;
	}
	\label{result-boxplot-ablation}
\end{figure*}

\subsubsection{Episode Visualization}
\label{subsec:3}
Taking the recovery scenario as an example, we visualize the process during which the multi-agent system handles faults of agents 2 and 3, respectively, to demonstrate our method's ability to manage communication maintenance and task allocation in an environment with potential agent faults.

As presented in Fig. \ref{recovery-figure}, three cooperative agents are set up in the lower left corner of the scene. Initially, all 3 agents run normally, maintaining formation and moving close to the prey. 
At time step 10, an agent breaks down. If agent 2 fails, since agent 3 does not have recovery capability, agents 1 and 2 jointly pursue the prey. To maintain communication, agent 1 does not continue to approach the prey quickly, but slows down and waits for agent 2 to enter the communication range, until time step 18. At this time, the prey chooses to escape to the left, and agents 1 and 2 predict the destination of escape in advance, successfully intercepting the prey at time step 26.

If agent 3 fails, since agent 2 has recovery capability, agent 2 first returns to the position of agent 3 to complete its recovery task,and agent 1 also turns back to maintain communication between them. Meanwhile, the prey chooses to stay slightly away from the boundary to ensure an open escape direction. At time step 24, agent 2 successfully completes the recovery task, and agents 2 and 3 continue the chase task.At time step 36, since the positions of agents 1 and 2 are biased to the right, the prey chooses to escape to the left. At time step 50, agents 1 and 2 complete the chase task.

\subsubsection{The Role of Attention}
\label{subsec:4}
\hfill\par
\textbf{Attention Distribution } 
To facilitate understanding of the attention distribution and its changes, we visualize the average attention distribution before and after the fault, as depicted in Fig. \ref{attention-figure}. Each row in the critic represents the attention allocation of an agent to three agents, detailedly row 1,2,3 for agent 1, agent 2, and agent 3. Each row in the actor represents the attention allocation of an agent to different pieces of information in its observation, detailedly row 1 for the token $e_{i0}$, row 2 for the agent itself, row 3 and row 4 for teammates, row 5 for the prey, and row 6 for obstacles.

When the system runs normally, the attention in the critic is relatively balanced compared to after the fault. In the actor, attention to the token is almost zero, as it has no practical significance. Attention to the prey is relatively high, with other information receiving varying attention. 

In the branch where agent 2 fails, agents 1 and 3 shift their attention from agent 2 to agent 1. In the branch where agent 3 fails, since agent 2 needs to recover it, both agents 1 and 2's critics pay more attention to the state of agent 2. Agent 2 focuses entirely on the location of the failed agent 3, while agent 1 primarily focuses on the location of the failed agent, while also considering the prey and other information. After completing the recovery task, agents are required to chase the prey again,so the attention weight quickly adjusts, with the critic shifting from agent 2 to agent 1 and the actor shifting from agent 3 to the prey. 

As can be seen, the distribution of attention is highly consistent with the process shown in the episode example in Fig. \ref{recovery-figure}, indicating that attention effectively guides the actor and critic in handling faults and other information.

\textbf{Ablation Study }
To better demonstrate the role of the attention module, we conducted ablation experiments. We compare AACFT with the following two variant methods: AAFT (removing attention from the critic in AACFT) and ACFT (removing attention from the actor in AACFT). As shown in Fig. \ref{result-curve-ablation} and Fig. \ref{result-boxplot-ablation}, AACFT performs better than AAFT and ACFT in most situations. The critic can guide the training of the actor, and the critic with attention is more accurate, as evidenced by ACFT's good performance in the relatively simpler abandonment scenario. However, AAFT shows better performance in the recovery scenario than in the abandonment scenario, indicating that the attention module in the actor is beneficial in complex scenarios requiring task redistribution and similar demands. The ablation study indicates that attention modules in both the actor and the critic are necessary.

\begin{figure}[h]
	\begin{center}
	\centering
	\includegraphics[width=0.85\linewidth]{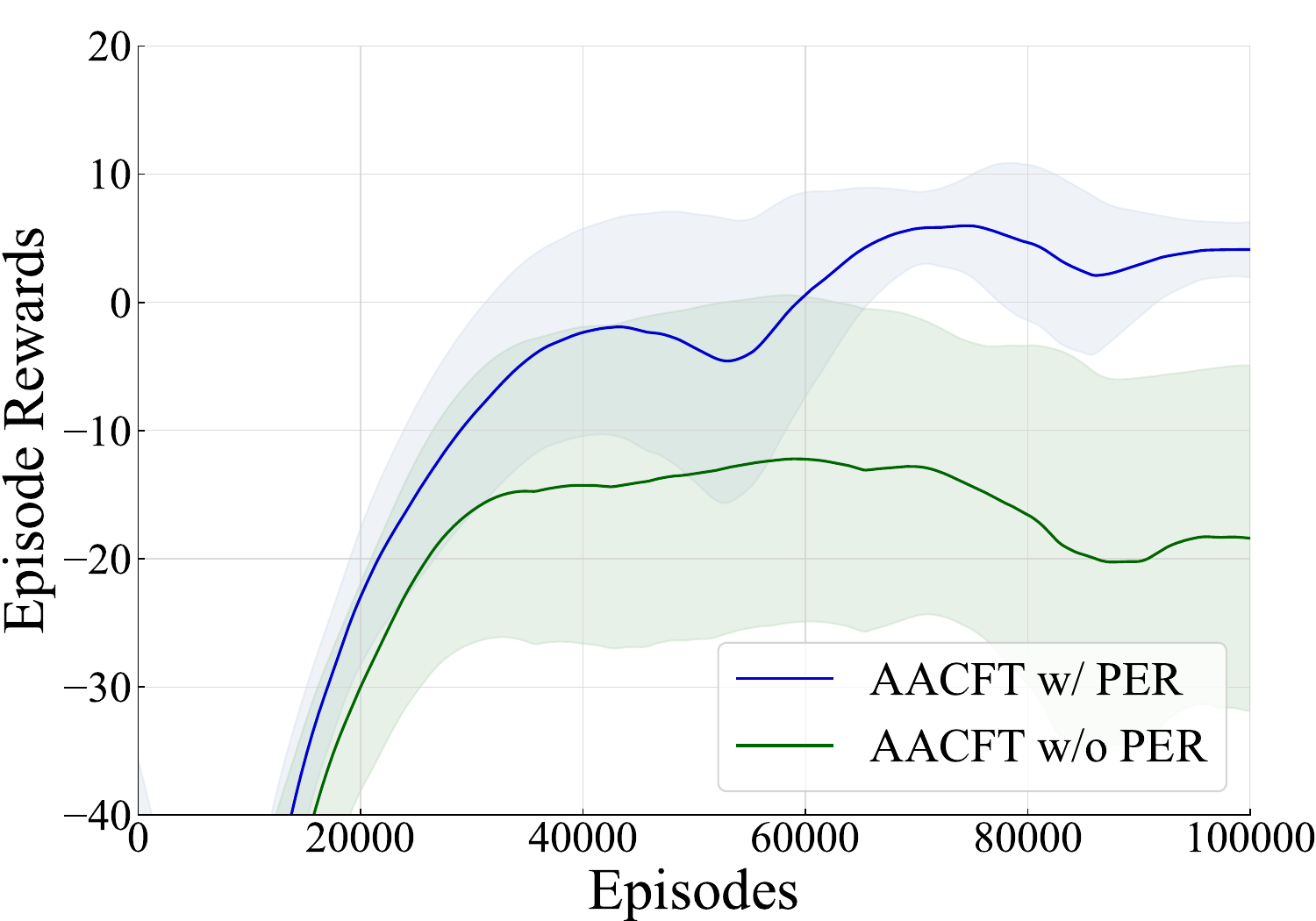}
	\caption{Reward curves from 4 training sessions for AACFT with and without PER in patrol scenario.}
	\label{result-curve-per}
	\end{center}
\end{figure}

\subsubsection{The Role of Priorty}
\label{subsec:5}
We select the more challenging patrol scenario to validate the role of PER in further enhancing fault tolerance performance.

As shown in Fig. \ref{result-curve-per}, in the patrol scenario, AACFT with PER performs significantly better than vanilla AACFT without PER. Vanilla AACFT struggles to learn any effective strategy, highlighting that the priority mechanism addresses the unique challenges posed by faults.

During training, the batch contains transitions of different types, with their quantities being a certain ratio. PER might bias towards particular types of transitions, causing a difference in the ratio compared to not using PER. To measure the magnitude of the difference introduced by PER, we define a new metric: Additional Sampling Rate. This metric indicates how much more a particular class of transitions is sampled with PER compared to without PER, given the same total number of transitions.

\begin{figure*}[h]
	\centering
	\subfloat[\label{fig:a}]{
		\includegraphics[width=0.4\linewidth]{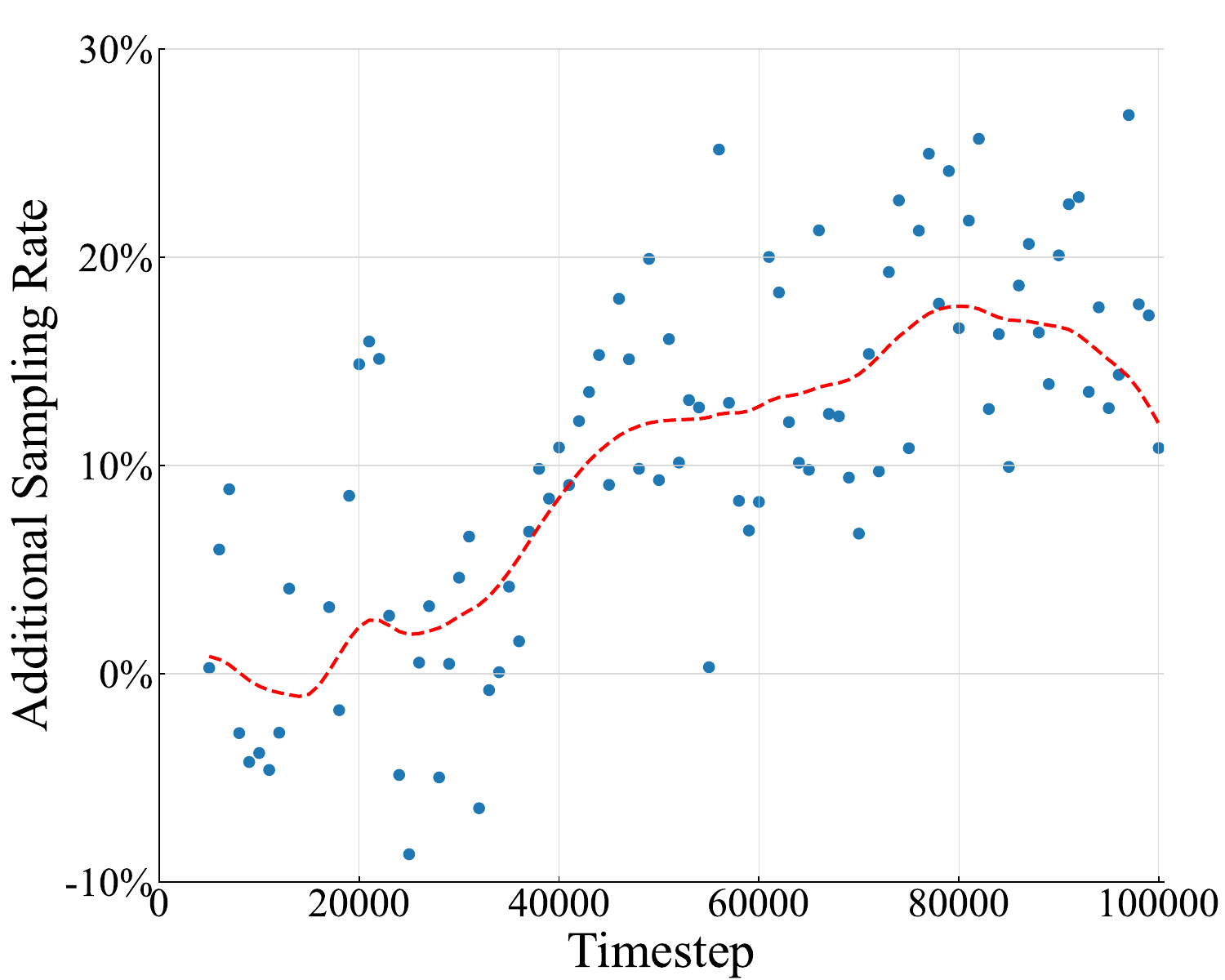}}
	\subfloat[\label{fig:b}]{
		\includegraphics[width=0.4\linewidth]{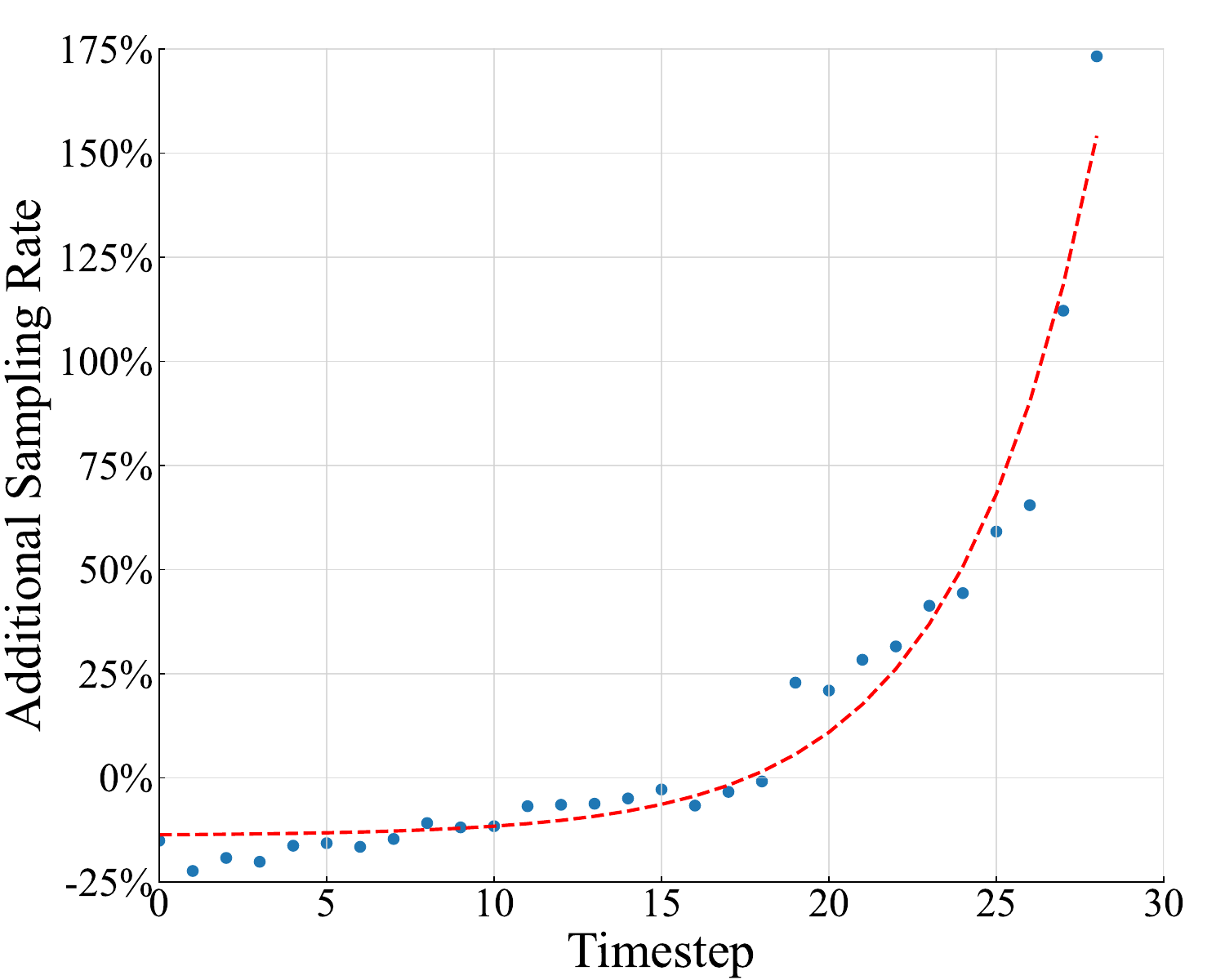}}
		\caption{Additional sampling rate in batches for training the critic for AACFT with PER in patrol scenario.
		(a) Additional sampling rate of post-fault transitions during training;
		(b) Additional sampling rate of pre-fault transitions at different time steps.
	}
	\label{visual-per}
\end{figure*}

\begin{equation}
	\text{Additional Sampling Rate} = \frac{N_{\text{PER}} - N_{\text{UNIF}}}{N_{\text{UNIF}}} \times 100\%,
\end{equation}
where $N_{\text{PER}}$ represents the number of a certain type's transitions using PER, while $N_{\text{UNIF}}$ denotes the number of this type's transitions under uniform sampling not using PER.

In Fig. \ref{visual-per}(a), we demonstrate the degree of PER's preference for post-fault transitions. It can be observed that during the first 40000 episodes, the Additional Sampling Rate fluctuates around 0, indicating no significant bias towards pre-fault or post-fault transitions. However, as the training progresses, its value increases to nearly 30\%, suggesting a tendency to train post-fault transitions once the pre-fault transitions have achieved satisfactory training results.

Fig. \ref{visual-per}(b) illustrates which timesteps PER is more inclined to select from among all pre-fault transitions. It can be observed that transitions from later time steps are significantly more likely to be sampled. This is because earlier transitions are often repeated, and once sufficiently trained, their training frequency can be reduced. The higher the time step, the closer the transitions are to the occurrence of the fault, indicating that PER better prepares AACFT for fault occurrences.

Through the above experiments, we demonstrate how PER improves training effectiveness on fault issues by preferentially selecting specific transitions.

\begin{table}[h]
	\caption{Performance on different time steps of fault.}
	\label{adaption-table}
	\begin{center}
	\begin{small}
	\begin{sc}
	\begin{tabular}{lccr}
	\toprule
	fault time & Completion rate\\
	\midrule
	5        & 0.789\\
	10       & 0.770\\
	15       & 0.798\\
	20       & 0.808\\
	25       & 0.846\\
	No fault & 0.841\\
	\bottomrule
	\end{tabular}
	\end{sc}
	\end{small}
	\end{center}
\end{table}

\subsubsection{Adaptability to Time}
\label{subsec:6}
The adaptability of algorithms to fault time is of concern due to the fact that the fault time of agents is often unpredictable in practice. We evaluate our AACFT model on different time steps of fault, each for 2000 episodes. As exhibited in \cref{adaption-table}, AACFT exhibits a high task completion rate, which increases with the delay of fault time, consistent with the intuitive expectation of a fault-tolerant system.

\section{Conclusion}
\label{conclusion}
In this paper, we propose a method that combines optimized model architecture with training data sampling strategy to address potential faults in MARL. Our algorithm leverages an attention mechanism to automatically identify faults and dynamically adjust the level of attention given to fault-related information. Simultaneously, a prioritization mechanism is employed to adjust the tendency of sample selection. These features allow the algorithm to automatically and efficiently utilize valuable data, effectively addressing the critical issues of fault tolerance.

Moreover, we have developed and open-sourced a new platform to support researchers in studying fault tolerance in MARL. Experimental results demonstrate that our method not only addresses the unique problems specific to MARL but also effectively deals with common issues such as communication maintenance and task allocation brought about by faults.

Looking ahead, we plan to design more types of faults within our platform to examine whether our method can adapt to the effects of different fault information and enhance its generalizability. Concurrently, we will explore the performance of the proposed method in more complex environments to further validate and enhance the method's effectiveness and practicality.

\bibliographystyle{IEEEtran}
\bibliography{liter}

\end{document}